%% file: sample-sigplan.tex
\documentclass[sigconf, nonacm, 9pt]{acmart}
\AtBeginDocument{%
  }

\acmYear{2025}\copyrightyear{2025}
\setcopyright{rightsretained}
\acmConference[Middleware '25]{26th International Middleware Conference}{December 15--19, 2025}{Nashville, USA}
\acmBooktitle{26th International Middleware Conference (Middleware '25), December 15--19, 2025, Nashville, USA}
\acmPrice{}
\acmDOI{10.1145/3590140.3629123}
\acmISBN{979-8-4007-0177-1/23/12}
\usepackage{algorithm}
\usepackage{algpseudocode}

\usepackage{amsmath, amssymb}
\usepackage[subtle]{savetrees}
\usepackage{multirow}
\usepackage{times}
\usepackage{latexsym}
\usepackage{booktabs}
\usepackage{graphicx}
\usepackage{amsmath}
\usepackage{algorithm}
\usepackage{algpseudocode}
\usepackage{tcolorbox}
\usepackage{graphicx}
\usepackage{caption}
\usepackage{subcaption}
\usepackage{multirow}
\usepackage{booktabs}
\usepackage{multirow}
\usepackage{color}
\usepackage{array}
\usepackage{geometry}
\usepackage{mathtools}
\usepackage{multicol}
\usepackage{tikz}
\usetikzlibrary{positioning,arrows.meta,calc}

\DeclareUnicodeCharacter{2212}{-}

\newcommand{\ours}{ShiftEx}
\begin{document}

\title{Shift Happens: Mixture of Experts based Continual Adaptation in Federated Learning}

\author{Rahul Atul Bhope}
\affiliation{%
  \institution{University of California, Irvine}
%  \streetaddress{1 Th{\o}rv{\"a}ld Circle}
  \city{Irvine, CA}
  \country{USA}
  }
%\email{rbhope@uci.edu}

\author{K. R. Jayaram}
\affiliation{%
  \institution{IBM Research AI}
%  \streetaddress{1 Th{\o}rv{\"a}ld Circle}
 \city{Yorktown Heights, NY}
  \country{USA}
  }
%\email{jayaramkr@us.ibm.com}

\author{Praveen Venkateswaran}
\affiliation{%
  \institution{IBM Research AI}
%  \streetaddress{1 Th{\o}rv{\"a}ld Circle}
 \city{Yorktown Heights, NY}
  \country{USA}
  }

\author{Nalini Venkatasubramanian}
\affiliation{%
  \institution{University of California, Irvine}
%  \streetaddress{1 Th{\o}rv{\"a}ld Circle}
  \city{Irvine, CA}
  \country{USA}
  }
%\email{nalini@ics.uci.edu}

%\email{gegi@us.ibm.com}



\input{abstract}
\keywords{federated learning, covariate shift, label shift, transfer learning, expert models, non-stationary data}

%\keywords{federated learning, covariate shift, label shift, transfer learning, expert models, non-stationary data}
\begin{comment}
    \begin{CCSXML}
<ccs2012>
<concept>
<concept_id>10010147.10010257.10010321</concept_id>
<concept_desc>Computing methodologies~Machine learning algorithms</concept_desc>
<concept_significance>500</concept_significance>
</concept>
<concept>
<concept_id>10010147.10010257.10010258.10010259.10010263</concept_id>
<concept_desc>Computing methodologies~Supervised learning by classification</concept_desc>
<concept_significance>500</concept_significance>
</concept>
<concept>
<concept_id>10010147.10010919.10010172</concept_id>
<concept_desc>Computing methodologies~Distributed algorithms</concept_desc>
<concept_significance>500</concept_significance>
</concept>
</ccs2012>
\end{CCSXML}

\ccsdesc[500]{Computing methodologies~Machine learning algorithms}
\ccsdesc[500]{Computing methodologies~Supervised learning by classification}
\ccsdesc[500]{Computing methodologies~Distributed algorithms}

\end{comment}



\maketitle
%\tableofcontents

\input{introduction}

\input{background}

\input{shiftex}

%\pvnote{What happens at the client, what happens at the server}

%\jkr{Add TEE for privacy preservation}

%\paragraph{Contrast with FedDrift.} 
%While FedDrift addresses general concept drift by clustering clients to the model with the lowest observed loss on their data, this approach is less suitable for covariate shift. Under covariate shift, the conditional distribution $P(y|x)$ remains stable while the input distribution $P(x)$ changes, meaning that an increase in loss may not necessarily indicate misalignment of the model's decision boundary but rather a domain gap in input features. In contrast, our method detects covariate shift explicitly using distributional distance metrics (e.g., MMD) and instantiates specialized expert models for clusters of clients with similar $P(x)$ shifts. This allows us to adapt to evolving input distributions while preserving decision consistency, avoiding unnecessary reassignment or retraining of well-calibrated models.

%\input{sysarchi}

\input{expertimental_strat.tex}

\input{results_updated.tex}

\input{results.tex}
\input{relatedworks.tex}

\input{conclusion}

%\input{table_compare}

%\begin{acks}
%We thank the anonymous reviewers and our shepherd, Dr. Ruben Mayer of MIDDLEWARE 2025, for their insightful feedback, which greatly improved this paper. We gratefully acknowledge the funding from the U.S. National Science Foundation (grant #2133391, grant #2008993) and DARPA (No. FA8750-16-2-0021). 
%\end{acks}

\bibliographystyle{ACM-Reference-Format}
\bibliography{sample-base}


\end{document}

%% file: abstract.tex
\begin{abstract}
Federated Learning (FL) enables collaborative model training across decentralized clients without sharing raw data, yet faces significant challenges in real-world settings where client data distributions evolve dynamically over time. This paper tackles the critical problem of covariate and label shifts in streaming FL environments, where non-stationary data distributions degrade model performance and necessitate a middleware layer that adapts FL to distributional shifts. We introduce \ours\, a shift-aware mixture of experts framework that dynamically creates and trains specialized global models in response to detected distribution shifts using Maximum Mean Discrepancy for covariate shifts. The framework employs a latent memory mechanism for expert reuse and implements facility location-based optimization to jointly minimize covariate mismatch, expert creation costs, and label imbalance. Through theoretical analysis and comprehensive experiments on benchmark datasets, we demonstrate 5.5-12.9 percentage point accuracy improvements and 22-95\% faster adaptation compared to state-of-the-art FL baselines across diverse shift scenarios. The proposed approach offers a scalable, privacy-preserving middleware solution for FL systems operating in non-stationary, real-world conditions while minimizing communication and computational overhead.

\end{abstract}

%% file: introduction.tex
\section{Introduction}

Federated Learning (FL) is increasingly being deployed in dynamic, real-world environments where 
data is not static but continuously generated by parties or clients over time. In such settings, models must 
be updated incrementally as new data arrives, a paradigm referred to as \textit{continual} or \textit{streaming} 
federated learning~\cite{kairouz2019advances, marfoq2023federatedlearningdatastreams,hamedi2025federatedcontinuallearningconcepts}. These evolving environments 
demand efficient mechanisms to handle incoming data, adapt models on the fly, and maintain performance 
without retraining from scratch.

To support these requirements, FL systems increasingly leverage techniques from 
traditional stream processing~\cite{armbrust2018structured, akidau2015dataflow}. A central 
technique is the use of \textit{windowing}, which segments an unbounded stream of data from parties into 
finite units or \textit{windows} that can be independently processed~\cite{akidau2015dataflow, lohrmann2015elastic}. 
This makes it feasible to apply traditional learning algorithms over streaming data. 
Among windowing strategies, the \textit{sliding window} is the most general form, where windows 
can overlap, shift, and vary in size. A common special case of the sliding window is 
the \textit{tumbling window}, which uses 
non-overlapping, fixed-size windows~\cite{zaharia2013discretized, lohrmann2015elastic}.
This allows for periodic updates to global models, facilitating continual adaptation to 
new information while preserving scalability and responsiveness.

\begin{figure}[thbp]
\centering
\includegraphics[width=\linewidth]{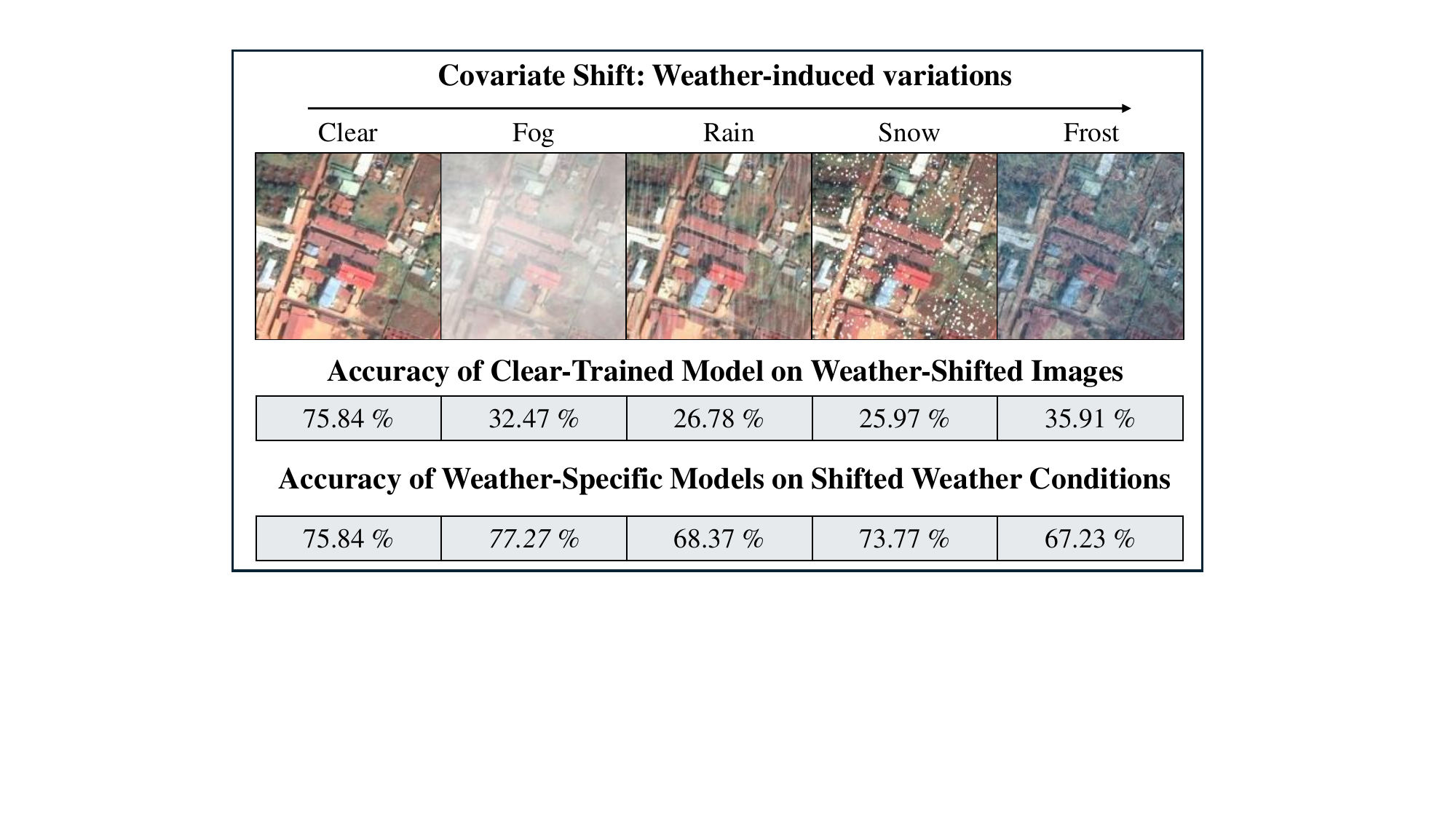}
\caption{Illustration of Covariate Shift: Weather-induced variations in visual input distributions simulate real-world covariate shifts impacting model generalization across different environmental conditions.}
\label{fig:covariate:shift}
\end{figure}

A critical challenge in streaming FL environments, is that the data from the parties will change and evolve over time. 
Consider the example shown in Figure \ref{fig:covariate:shift} depicting satellite imagery to study land-use patterns. We observe differences in the images based on the changing weather conditions (e.g.) image clarity deteriorates in the winter when there is snow and frost. 
Hence, as new data arrives, both the input distribution $P(X)$ and label distribution $P(Y)$ can shift ~\cite{zhao2018federated}. These distributional shifts, known as \emph{covariate shift}
and \emph{label shift} respectively, pose significant challenges for traditional FL approaches 
that maintain a single global model~\cite{yoon2021fedmix}.

As shown in Figure~\ref{fig:covariate:shift}, when parties experience different distribution shifts at different times, aggregating their 
model updates can lead to performance degradation and negative knowledge transfer~\cite{li2021fedbn}. 
The global model fails to adapt effectively to the diverse and evolving data patterns across the 
federation, resulting in poor generalization for parties whose distributions have shifted.
In contrast, specialized \textit{expert} models, trained on each specific shifted data distributions significantly outperform the global training strategy.

From a middleware systems perspective, streaming FL environments present fundamental challenges in distributed data processing and service orchestration, particularly when dealing with evolving data distributions and shifts that require dynamic system reconfiguration.  Inspired by the effectiveness of \textit{expert} models, and to address these challenges, this paper introduces a comprehensive framework for handling both covariate and label shifts in streaming FL using a Mixture of Experts (MoE)~\cite{6797059}-based approach. Each expert model specializes in a distinct covariate regime and is dynamically adapted to evolving label distributions across windows. Our framework detects shifts in real-time, assigns parties to appropriate experts, and mitigates the risk of catastrophic forgetting during adaptation. Our key contributions include:

%\rbnote{to revisit}
\begin{itemize}
    \item An online shift detection mechanism that identifies both covariate and label shifts in real-time, enabling shift-aware routing of parties without requiring aggregator-side access to labels.
    %\item We propose an online shift detection mechanism that leverages Maximum Mean Discrepancy (MMD) for identifying covariate shift and Jensen–Shannon Divergence (JSD) for detecting label shift, enabling shift-aware routing decisions without requiring labels on the server.
    \item A dynamic expert management strategy that creates, reuses, and adapts specialized models based on detected shifts and clustering patterns on the parties.
    %\item We introduce a dynamic expert instantiation and reuse strategy that creates or reassigns expert models to client clusters based on the nature and degree of detected shifts, allowing for scalable and flexible specialization.
    \item A formalized expert assignment framework that optimally matches parties to experts while balancing specialization benefits and system efficiency.
    %\item We formalize the expert assignment process as a constrained optimization problem that jointly minimizes covariate mismatch, adaptation cost, and expert imbalance, while maximizing expected utility under shifting distributions.
\item Experimental validation showing \textbf{5.5--12.9\% point} accuracy improvements and \textbf{22--95\%} faster adaptation compared to state-of-the-art FL baselines across diverse datasets with realistic distribution shifts.
%\item Through extensive experiments on diverse datasets exhibiting both real-world and synthetic distributional shifts, we demonstrate that our framework consistently outperforms state-of-the-art FL baselines by \textbf{6.5--13.8\%} in maximum accuracy, while reducing adaptation time (i.e., communication rounds to recover) by \textbf{80--100\%}, showcasing significant improvements in accuracy, convergence stability, and adaptability to drift.

\end{itemize}

The remainder of this paper is organized as follows: Section 2 provides background and challenges in streaming FL and distribution shifts. Section 3 details our shift-aware expert framework. Sections 4 and 5 detail the part and the aggregator side algorithm. Section 6 depicts our Experimental Strategy, Section 7 discusses the results, Section 8 discusses related work in this field, and Section 9 concludes with a discussion of limitations and future work. In the rest of this paper, we shall use the words ``client'' or ``party'' to refer to a participant in an FL job, and ``server'' or ``aggregator'' to refer to the coordinator of the FL job.

%The remainder of this paper is organized as follows: Section 2 provides a background to the problem. Section 3 details our strategy for handling shifts in FL with streaming data and provides a mathematical formulation of our approach. Section 4 introduces our shift-aware expert instantiation framework. Section 5 presents experimental results, and Section 6 concludes with a discussion of limitations and future work.

%% file: background.tex
\section{Background and Challenges}

Federated Learning (FL) enables decentralized model training across multiple parties while preserving data privacy. Unlike centralized learning, parties train models locally on private data and share only model updates with a central aggregator. While federated learning has been designed to be privacy-preserving, several attack vectors have been identified, including model inversion attacks that attempt to reconstruct training data from gradients~\cite{zhu2019deep}, membership inference attacks that determine if specific samples were used in training~\cite{melis2019exploiting}, and poisoning attacks that inject malicious updates to degrade model performance~\cite{bagdasaryan2020backdoor}. However, extensive research has developed robust defenses against these vulnerabilities, including differential privacy mechanisms~\cite{abadi2016deep}, secure aggregation protocols~\cite{bonawitz2017practical}, gradient compression techniques~\cite{seide2014feature}, and Byzantine-robust aggregation methods~\cite{blanchard2017machine}, effectively mitigating most, if not all, of these attack vectors in practice. 

Nevertheless, FL systems deployed in real-world settings face a critical challenge that remains largely unaddressed: party data distributions evolve over time, creating dynamic learning environments that traditional FL approaches struggle to handle effectively.

%Federated Learning (FL) enables decentralized model training across multiple parties while preserving data privacy. Unlike centralized learning, parties train models locally on private data and share only model updates with a central aggregator. However, FL systems deployed in real-world settings face a critical challenge: party data distributions evolve over time, creating dynamic learning environments that traditional FL approaches struggle to handle effectively. \rbnote{Missing: Privacy, mitigation strategies on privacy attacks}

\subsection{Distribution Changes in Streaming FL}

%\pvnote{we tackle both, as long as there is a change in distribution}

%\pvnote{notion of shift depends on the reference, technically it could be the first window, user can specify the reference, the approach would work for both}

In windowed streaming FL, distributional changes manifest as differences between consecutive time windows. These changes can occur abruptly, where a party's data distribution shifts dramatically between adjacent windows, or gradually through accumulated changes across multiple windows. Both scenarios pose unique challenges for federated systems, where model aggregation assumes some degree of distributional consistency across participants. \textbf{Shift} refers to abrupt, discontinuous changes in the data distribution, often occurring suddenly between consecutive windows and disrupting model performance without warning. Such shifts can be triggered by events like sensor recalibration or sudden changes in class prevalence, and are typically geographically localized and immediate, affecting specific parties or subsets of the FL deployment in a noticeable way \cite{quinonero2009dataset}. 

In contrast, \textbf{drift} characterizes gradual, continuous changes in data distribution or the underlying data-generating process, unfolding over longer timescales. Examples include seasonal changes in user behavior or progressive wear in hardware sensors. Rather than being a one-time perturbation, drift consists of a sequence of small shifts that accumulate and degrade model performance over time. Unlike sudden shifts, drift is systemic and harder to detect, often requiring sustained monitoring and adaptation to maintain performance \cite{gama2014survey}. One effective strategy to mitigate the impact of drift is to adapt the model incrementally as new data arrive, before the accumulation of shifts becomes disruptive. This continual adaptation helps maintain alignment between the model and the evolving data distribution, thereby improving robustness to long-term drift \cite{lu2018learning}.

The federated setting amplifies these challenges in several ways. First, the aggregator cannot directly observe party data to detect shifts or drifts, requiring detection mechanisms on the aggregator side that rely on model updates or limited statistics. Second, parties may experience shifts at different times and rates, making synchronized adaptation difficult. Third, aggregating updates from parties with divergent distributions can lead to negative knowledge transfer, where learning from shifted parties degrades performance on stable ones.

While shift and drift differ in their temporal characteristics and detectability patterns, they fundamentally represent the same underlying challenge: distributional changes that degrade federated model performance. Our framework addresses both types of distributional changes through unified detection and adaptation mechanisms. For clarity and simplicity, throughout the remainder of this paper, we use the term ``shift'' to refer to both abrupt shifts and gradual drift, unless the distinction is specifically relevant to the discussion.

%\rbnote{We tackle both shift and drift scenarios, as long as there is a distribution change—irrespective of how it evolves. We tackle both shift and drift. In the rest of the paper we use shift and drift interchangeably}%The notion of shift itself depends on the reference: it could be the initial training window, or any user-specified baseline. Our approach remains valid regardless of how the reference is defined, enabling flexibility in deployment.}

\subsection{Types of Distributional Changes}

{\bf Covariate Shift} occurs when the input distribution $P(x)$ changes while the conditional relationship $P(y|x)$ remains stable. This is common in audio-visual domains where noise and environmental factors affect the appearance of inputs, but not their semantic meaning. For example, an image classifier trained on clear weather images may encounter foggy or rainy conditions, changing the visual appearance without altering the underlying object categories. In federated settings, different parties may experience covariate shifts due to varying local conditions, device characteristics, or environmental factors. If not addressed, this change in $P(x)$ can lead to reduced model performance despite the label semantics remaining stable.

{\bf Label Shift} involves changes in the marginal label distribution $P(y)$ while keeping $P(x|y)$ fixed. This commonly occurs in applications like land use classification from satellite imagery, where seasonal or geographic factors change the prevalence of different land types without altering their visual characteristics. In FL systems, label shift can occur when party populations change, for example, a healthcare federation where disease prevalence varies by region or season.

{\bf Concept Shift} represents changes in the conditional distribution $P(y|x)$, where the fundamental relationship between inputs and labels evolves. While this is the most challenging form of distributional change, it typically requires labeled data or domain expertise to detect and adapt to, making it less tractable in privacy-preserving FL settings.

\textbf{Impact on Traditional FL}. Traditional FL algorithms (\textit{FedAvg~\cite{mcmahan2017communication}, FedProx~\cite{fedprox}}), which aggregate model updates into a single global model, struggle in such heterogeneous settings. For covariate shifts, the global model may not generalize to parties with divergent feature distributions. For label shifts, optimization is biased toward dominant labels, degrading performance on less-seen classes. As a result, models trained under such aggregation schemes tend to be brittle and fail to generalize across the diverse distributions encountered during deployment.

%\rbnote{add moe, how it is a challenge totrain moe in FL, because you'll need a gating netowkr}

%\rbnote{rephriase routing to assignment}

\subsection{The Federated MoE Challenge}

A natural way to handle shifts is to allow multiple models to coexist, each specializing in different distributional regimes. This is precisely the insight behind {\bf Mixture of Experts (MoE)}, a machine learning paradigm where multiple "expert" models are trained to handle different aspects or subsets of the data space. This idea has led to the development of centralized MoE architectures~\cite{masoudnia2014mixture,cai2024survey}. They typically consist of a gating network that is jointly trained with a fixed set of expert models using the full visibility of the global data distribution. This gating function learns to map individual inputs to the appropriate experts by observing how different experts perform on different data or task subsets.

However, this approach presents fundamental challenges in federated learning and evolving data distribution settings for several reasons. First, due to privacy constraints, the aggregator cannot access data from parties to train the centralized gating function. 
Second, a fixed set of experts may be insufficient to handle new or unseen covariate regimes, requiring an architecture that can dynamically train new experts on demand.
Third, the participation of individual parties is typically sparse and asynchronous -- not all parties are available at every training round-- presenting communication challenges, and also making it difficult to learn stable routing patterns.

Consequently, we need a method to train a group of expert models in a federated manner, such that data never leaves any party. Also, for privacy preservation at inference, instead of routing data to experts for inference, a party should be associated with an expert model, whose parameters are transmitted to the party
for local inference.

%\subsection{Our Focus and Motivation}

%This work focuses on covariate and label shifts for several practical reasons. Both can be detected using techniques that preserve FL's privacy constraints—covariate shift through unsupervised methods applied to party model updates or feature statistics, and label shift through party-side label proportion estimation. Additionally, these shifts are well-suited to windowed detection mechanisms, as they often manifest as observable changes between consecutive time windows.

%More importantly, covariate and label shifts represent the most common distributional changes in real FL deployments. They can be addressed through specialization strategies where different expert models handle different distributional regimes, providing a natural motivation for mixture-of-experts approaches. This specialization allows the federation to maintain multiple models adapted to different party conditions while avoiding the negative knowledge transfer that occurs when aggregating updates from distributionally diverse parties.

%The combination of detectability, prevalence, and amenability to expert-based solutions makes covariate and label shifts the ideal focus for developing practical shift-aware FL systems that can operate effectively in streaming environments.

%% file: shiftex.tex
\section{\ours: Shift-Aware Mixture of Experts Framework}

This work focuses on covariate and label shifts for several practical reasons. Both can be detected using techniques that preserve FL's privacy constraints—covariate shift through unsupervised methods applied to party model updates or feature statistics, and label shift through party-side label proportion estimation. Additionally, these shifts are well-suited to windowed detection mechanisms, as they often manifest as observable changes between consecutive time windows.

More importantly, covariate and label shifts represent the most common distributional changes in real FL deployments. They can be addressed through specialization strategies where different expert models handle different distributional regimes, providing a natural motivation for mixture-of-experts approaches. This specialization allows the federation to maintain multiple models adapted to different party conditions while avoiding the negative knowledge transfer that occurs when aggregating updates from distributionally diverse parties.

The combination of detectability, prevalence, and amenability to expert-based solutions makes covariate and label shifts the ideal focus for developing practical shift-aware FL systems that can operate effectively in streaming environments.

\subsection{Our Approach: Party-Level Expert Assignment}

We propose \ours, a shift-aware federated learning framework that sidesteps the gating network challenge by operating at the party level rather than the input level. Instead of routing individual inputs to experts, our system assigns experts to parties based on distributional similarity and detected shifts.
This party-to-expert assignment approach offers several advantages. First, it preserves privacy since assignment decisions are made using aggregate statistics or model-level information rather than raw data. Second, it reduces communication overhead by maintaining stable party-expert associations over time windows rather than per-input routing decisions. Third, it naturally aligns with the federated training process, where model updates are already aggregated at the party level.

%\rbnote{merge with 3.4}

%\rbnote{remove instantiation, say experts models are trained}

\ours\ treats covariate and label shifts as complementary signals for expert management. When covariate shift is detected, the system determines whether existing experts can handle the new input distribution or if new expert specialization is needed. When label shift occurs, the system rebalances participant experts to ensure experts maintain diverse and representative training data across label categories.
The framework unifies these responses through coordinated expert assignment and participant rebalancing: expert models are created and specialized to handle distinct input regimes (addressing covariate shift), while participant assignment strategies ensure balanced learning across label distributions (addressing label shift). This dual approach enables robust adaptation to both types of distributional change while maintaining the scalability and privacy benefits of FL.

\subsection{\ours\ Architecture}

\begin{figure}[htbp]
\centering
\includegraphics[width=\linewidth]{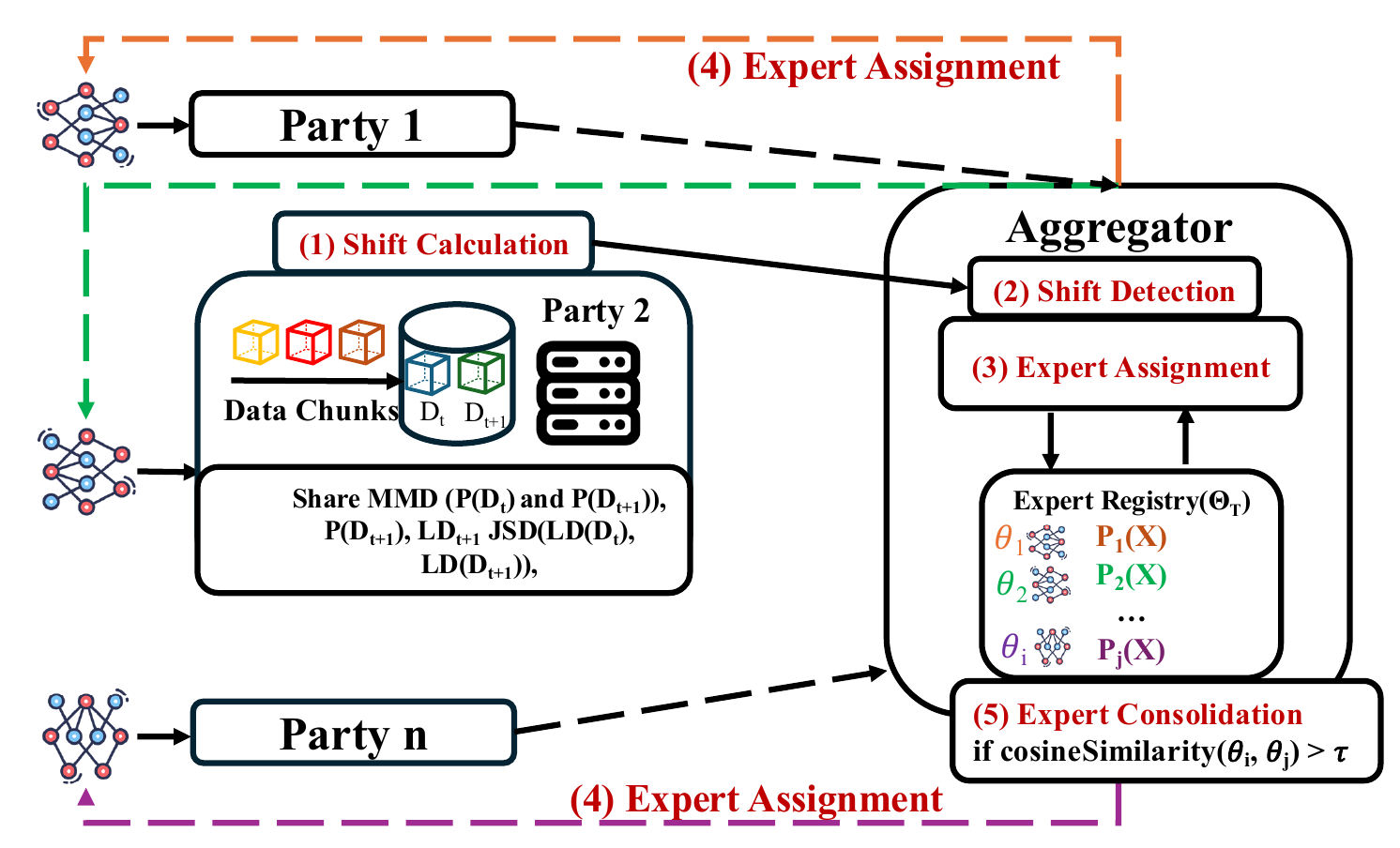}
\caption{System overview of \ours\ for FL.}
\label{fig:shifted:combined}
\end{figure}

\ours\ differs from traditional continual federated learning (FL) systems by explicitly targeting both covariate and label shifts using an MMD-based shift detector and a distribution-aware Mixture-of-Experts (MoE) model assignment strategy. The system is designed to be modular and can be integrated into existing FL frameworks such as PySyft~\cite{ziller2021pysyft} or Flower~\cite{beutel2020flower}.

Each client maintains a local dataset and runs a stream processing engine (e.g., Apache Kafka~\cite{garg2013apache} or Flink~\cite{carbone2015apache}) to collect, ingest, and preprocess incoming data streams, storing them in a local database. This setup resembles standard streaming data platforms. \ours\ extends this setup by incorporating a shift detection module that computes latent feature representations and label distributions from local data across consecutive windows. Covariate shift is identified using Maximum Mean Discrepancy (MMD) on latent representations, while label shift is detected via Jensen–Shannon Divergence (JSD) on label histograms. After computing these statistics, clients transmit only the aggregated embeddings, label distributions, and corresponding MMD and JSD scores to the aggregator. MMD and JSD were selected because they are non-parametric and lightweight, introducing minimal overhead; however, the framework itself is detector-agnostic and can readily accommodate alternative choices if desired.

On the aggregator side, \ours\ first uses a set of precomputed thresholds to detect covariate and label shift and maintains a registry of expert models, each tagged with its corresponding covariate regime, represented by aggregated latent embeddings from past windows. At window $t=0$, the registry contains only a single expert. In subsequent windows,
the aggregator uses the MMD scores to match each party (or group of parties) to the most appropriate expert. Specifically, parties are first clustered based on similarity in their latent feature distributions. Each cluster is then (1) assigned to an existing expert from the registry if the MMD between the cluster’s aggregated representation and the expert's profile is below a threshold, or (2) set up to train a new expert model using a FL process among its members if no suitable match is found. For training, we employ FLIPS~\cite{10.1145/3590140.3629123}, which further clusters parties based on their label histograms to promote class-balanced training.

A key distinction from centralized MoE systems is that expert assignment in \ours\ occurs at the client level rather than at the granularity of individual inputs. Each client is associated with the most suitable expert (or, if necessary, a newly created one) based on latent statistics derived from its local data distribution. At inference time, the process is straightforward: the client simply uses the parameters of the expert model it was previously assigned and locally trained with. This avoids the need for a centralized gating network, ensures that inference is always well defined, preserves privacy by preventing raw inputs from being routed to a server, and reduces overhead by eliminating per-sample routing decisions.

%This ensures that, can be directly applied here to improve fairness and stability during model updates.

%After expert assignment, parties within the same cluster—i.e., those with similar covariate characteristics, share the same expert model. To ensure balanced label distributions within each expert group, we integrate a label stratification mechanism. Techniques such as 

Overall, this architecture as seen in Figure~\ref{fig:shifted:combined}, allows \ours\ to be flexibly deployed on top of any existing FL framework, enabling robust, adaptive training in the presence of streaming data shifts while preserving modularity, scalability, and compatibility with existing systems.

\section{\ours\ --  Party Side}

\begin{algorithm}
\caption{Party-Side: Shift Detection}
\label{alg:party}
\begin{algorithmic}[1]
\small
\Require 
    Current local dataset $\mathcal{D}_t^c = \{(x_i, y_i)\}$, 
    Previous Dataset $\mathcal{D}_{t-1}^c$, 
    Thresholds $\delta_{\text{cov}}, \delta_{\text{label}}$
\Ensure 
    Covariate profile $P_t^c(X)$, label histogram $\mathbf{y}_t^c$, covariate shift $\Delta_{\text{cov}}$, label shift $\Delta_{\text{label}}$

\State Compute feature embeddings $\phi(x_i)$ for all $x_i \in \mathcal{D}_t^c$
\State Estimate $P_t^c(X)$ as the distribution over embeddings
\State Compute normalized label histogram $\mathbf{y}_t^c$
\If{$\mathcal{D}_{t-1}^c$ exists}
    \State $\Delta_{\text{cov}} \gets \mathrm{MMD}(P_t^c(X), P_{t-1}^c(X))$
    \State $\Delta_{\text{label}} \gets \mathrm{JSD}(\mathbf{y}_t^c, \mathbf{y}_{t-1}^c)$
\Else
    \State $\Delta_{\text{cov}} \gets 0$, $\Delta_{\text{label}} \gets 0$
\EndIf
\State Transmit $\{P_t^c(X), \mathbf{y}_t^c, \Delta_{\text{cov}}, \Delta_{\text{label}}\}$ to the aggregator
\end{algorithmic}
\end{algorithm}

%\subsection{Initial Setup and Windowing}

Each participant processes its streaming data using windows of fixed duration. These windows can be of tumbling or sliding in nature, segmenting the continuous data stream into time periods. The system supports both tumbling and sliding window configurations, with the system responding when the shift is greater than a predefined threshold. This windowing strategy ensures synchronized training cycles across all parties while enabling periodic model updates as new data arrives. Within each window, parties accumulate training data that will be used for local model training and subsequent aggregation.

\subsection{Bootstrap Phase with FLIPS}~\label{bootstrap-flips}

At system initialization, no global model exists, requiring all parties to collaborate in training an initial baseline model. 
To ensure representative and balanced training, \ours\ employs FLIPS (Federated Learning with Intelligent Participant Selection) 
for participant selection during this bootstrap phase~\cite{10.1145/3590140.3629123}. FLIPS performs label distribution clustering by 
analyzing the label distribution of each party's data a priori, grouping parties with similar label proportions into clusters. 
During federated training rounds, FLIPS ensures that participants are selected such that each cluster is equitably represented, 
preventing bias toward parties with overrepresented label distributions and promoting a more robust initial global model.

This label-aware participant selection is particularly crucial for \ours\ because the quality of the initial global model affects subsequent shift detection and expert assignment decisions. By ensuring the bootstrap model is trained on representative data across different label distributions, FLIPS provides a stable foundation for detecting when parties begin to experience distributional shifts that warrant expert specialization. The clustering information from FLIPS also provides valuable baseline statistics about party label distributions, which \ours\ later uses to detect label shift and inform party-to-expert assignment decisions.

\subsection{Covariate Shift Detection via Maximum Mean Discrepancy}

One of the objectives of \ours\ is to identify the parties that have undergone covariate and/or label shifts at a 
given time instant $t$. After the conclusion of a federated learning round through FLIPS as described above, 
parties process their local datasets $\mathcal{D}_t^c = \{(x_i, y_i)\}$ through their current model 
at timestep $t$ to obtain feature embeddings: $P^c_t(x_i) = f_{\theta}(x_i)$

These embeddings compress raw inputs into a compact, task‑relevant feature space. If this latent representation 
changes substantially from one window to the next, the parties can ascertain that the \emph{underlying distribution} has drifted -- 
even if they never look at the raw pixels or tokens themselves. This makes the covariate shift detection 
model-, data-, and label-agnostic because raw data can always be translated into a latent representation.

To assess changes in the input distribution, parties compute the Maximum Mean Discrepancy (MMD) between the current and previous window's embeddings: $\mathrm{MMD}(P_t^c(X),  P_{t-1}^c(X))$. MMD is a non-parametric statistical measure that compares two probability distributions by embedding them into a reproducing kernel Hilbert space (RKHS) and computing the distance between their means in that space~\cite{JMLR:v13:gretton12a}. 

Formally, given two distributions $P$ and $Q$ with samples $\{x_i\}_{i=1}^{n_P}$ and $\{y_j\}_{j=1}^{n_Q}$, MMD is defined as:
\begin{align}
\mathrm{MMD}^2(P, Q) &= \mathbb{E}_{x,x' \sim P}[k(x,x')] \\
&\quad + \mathbb{E}_{y,y' \sim Q}[k(y,y')] \nonumber \\
&\quad - 2\mathbb{E}_{x \sim P, y \sim Q}[k(x,y)] \nonumber
\end{align}
where $k(\cdot, \cdot)$ is a positive definite kernel function (we use the RBF kernel $k(x,y) = \exp(-\gamma \|x-y\|^2)$).

The key advantage of MMD is that it provides a principled way to detect distributional differences without requiring distributional assumptions or labeled data. It operates directly on feature representations, making it ideal for privacy-preserving covariate shift detection in federated settings. A high MMD value ($> \delta_{\text{cov}}$) indicates a significant change in the feature distribution, signaling covariate shift. 

\subsection{Label Shift Detection via Jensen-Shannon Divergence:} To detect label distribution changes, we employ Jensen-Shannon Divergence (JSD)~\cite{menendez1997jensen}, a symmetric and bounded measure for comparing discrete probability distributions. Unlike the asymmetric Kullback-Leibler divergence~\cite{kullback1951kullback}, JSD is symmetric ($\mathrm{JSD}(P||Q) = \mathrm{JSD}(Q||P)$) and always finite, making it robust for comparing label distributions that may have non-overlapping support.

For two discrete distributions $P$ and $Q$, JSD is defined as:
\[
\mathrm{JSD}(P||Q) = \frac{1}{2}D_{KL}(P||M) + \frac{1}{2}D_{KL}(Q||M)
\]

where $M = \frac{1}{2}(P + Q)$ is the average distribution, and $D_{KL}$ denotes Kullback-Leibler divergence. In our context, we compute JSD between normalized label histograms $\hat{\mathbf{y}}_t^c$ and $\hat{\mathbf{y}}_{t-1}^c$, where $\hat{\mathbf{y}}_t^c[i] = \frac{|\{y \in \mathcal{D}_t^c : y = i\}|}{|\mathcal{D}_t^c|}$ represents the proportion of samples belonging to class $i$ in window $t$. JSD values range from 0 (identical distributions) to $\log(2)$ (completely disjoint distributions), providing an interpretable measure of label distribution shift.

To summarize, at each time window $t$, parties compute local statistics to detect shift:
\begin{itemize}
    \item Covariate shift: $\mathrm{MMD}(P_t^c(X), P_{t-1}^c(X)) > \delta_{\text{cov}}$
    \item Label shift: $\mathrm{JSD}(\hat{\mathbf{y}}_t^c, \hat{\mathbf{y}}_{t-1}^c) > \delta_{\text{label}}$
\end{itemize}

Parties transmit the computed statistics $\{\mathrm{MMD}_t^c, \mathrm{JSD}_t^c, P_t^c(X), \hat{\mathbf{y}}_t^c\}$ to the aggregator, enabling privacy-preserving shift detection without sharing raw data.

\section{\ours\ - Aggregator Side Operations}

\begin{algorithm}
\caption{Aggregator-Side: Shift-Aware Federated Learning with Expert Assignment}
\label{alg:aggregator}
\begin{algorithmic}[1]
\small
\Require 
    Initial model $\theta_0$, party set $\mathcal{C}$, 
    thresholds $\delta_{\text{cov}}, \delta_{\text{label}}, \tau, \epsilon$, minimum cluster size $\gamma$
\Ensure 
    Final expert pool $\Theta_T = \{\theta_T^{(1)}, \dots, \theta_T^{(K)}\}$

% ---- INITIALIZATION PHASE ----
\State \textcolor{blue}{\tiny // Initialize expert pool and assign to all parties}
\State Initialize expert pool $\Theta_0 \gets \{\theta_0\}$; assign $M_0(c) \gets \theta_0$ for all $c \in \mathcal{C}$

\For{each round $t = 1$ to $T$}
    \State \textcolor{blue}{\tiny // Receive latent and label statistics from all parties}
    \State Receive $\{P_t^c(X), \mathbf{y}_t^c, P_{t-1}^c(X), \mathbf{y}_{t-1}^c\}$ from all $c \in \mathcal{C}$

    \State \textcolor{blue}{\tiny // Perform shift detection using thresholds}
    \State $\mathcal{C}^{\text{shift}} \gets \{c \in \mathcal{C} \mid \mathrm{MMD}(P_t^c(X), P_{t-1}^c(X)) > \delta_{\text{cov}}$ or $\mathrm{JSD}(\mathbf{y}_t^c, \mathbf{y}_{t-1}^c) > \delta_{\text{label}}\}$

    \If{$\mathcal{C}^{\text{shift}} \neq \emptyset$}
        \State \textcolor{blue}{\tiny // Cluster shifted parties using latent representations}
        \State $\{G_1, \dots, G_m\} \gets \textsc{KMeans}(\mathcal{C}^{\text{shift}}, P_t^c(X))$

        \ForAll{groups $G_j$}
            \If{$|G_j| \geq \gamma$}
                \State \textcolor{blue}{\tiny // Match to existing expert or create a new one}
                \State Aggregate $P_j(X) \gets \bigcup_{c \in G_j} P_t^c(X)$
                
                \If{\textsc{MatchExpert}($P_j(X), \Theta_{t-1}$, $\epsilon$)} 
                    \State $\theta_k \gets$ closest matching expert
                    \State $\mathcal{C}_j^{\text{balanced}} \gets \textsc{FLIPS}(G_j)$ 
                    \State $\theta_k \gets \textsc{FL}(\mathcal{C}_j^{\text{balanced}}, \theta_k)$ \textcolor{blue}{\tiny // Label-balanced training}
                    
                \Else
                    \State $\theta_{\text{new}} \gets \textsc{Clone}(\theta_0)$
                    \State $\mathcal{C}_j^{\text{balanced}} \gets \textsc{FLIPS}(G_j, y_c^t)$
                    \State $\textsc{SAVE}(\theta_k, P_j(X))$
                    \State $\theta_{\text{new}} \gets \textsc{FL}(\mathcal{C}_j^{\text{balanced}}, \theta_{\text{new}})$
                    \State $\Theta_t \gets \Theta_{t-1} \cup \{\theta_{\text{new}}\}$
                    \State Assign $M_t(c) \gets \theta_{\text{new}}$ for $c \in G_j$
                \EndIf

            \Else
                \State \textcolor{blue}{\tiny //Cluster too small – trigger local finetuning}
                \State \textsc{LocalFineTune}$(G_j, M_{t-1}, M_t)$
            \EndIf
        \EndFor
    \EndIf

    \State \textcolor{blue}{\tiny //Merge experts with high similarity to prevent redundancy}
    \ForAll{pairs $(\theta_i, \theta_j) \in \Theta_t$}
        \If{$\textsc{ModelSimilarity}(\theta_i, \theta_j) > \tau$}
            \State $\theta_{\text{merged}} \gets \textsc{ConsolidateExperts}(\theta_i, \theta_j)$
            \State $\Theta_t \gets (\Theta_t \setminus \{\theta_i, \theta_j\}) \cup \{\theta_{\text{merged}}\}$
            \State Update $M_t$ for affected parties
        \EndIf
    \EndFor
\EndFor

\State \Return $\Theta_T$
\end{algorithmic}
\end{algorithm}

After the bootstrap phase (Section~\ref{bootstrap-flips}), the aggregator maintains the initial global model as the first expert and begins processing party statistics to detect and respond to distribution shifts. At each window $t$, the aggregator receives shift statistics $\{\mathrm{MMD}_t^c, \mathrm{JSD}_t^c, P_t^c(X), \hat{\mathbf{y}}_t^c\}$ from participating parties and orchestrates expert assignment and training.

The aggregator first identifies parties experiencing distributional shifts by comparing their reported statistics against predefined thresholds:
\begin{itemize}
\item \textbf{Covariate shift parties}: $\mathcal{C}^{\text{cov}} = \{c : \mathrm{MMD}_t^c > \delta_{\text{cov}}\}$  
\item \textbf{Label shift parties}: $\mathcal{C}^{\text{label}} = \{c : \mathrm{JSD}_t^c > \delta_{\text{label}}\}$
\item \textbf{Stable parties}: $\mathcal{C}^{\text{stable}} = \mathcal{C} \setminus (\mathcal{C}^{\text{cov}} \cup \mathcal{C}^{\text{label}})$
\end{itemize}

The thresholds $\delta_{\text{cov}}$ and $\delta_{\text{label}}$ are derived during the bootstrap phase from the null distributions of MMD and JSD scores. $\delta_{\text{cov}}$ is set via p-value estimation from bootstrapped client feature representations assuming no shift, while $\delta_{\text{label}}$ is based on JSD statistics between predicted and prior label distributions under stable conditions.

\subsection{Optimal Expert Assignment}

After a window-level shift is detected, the aggregator must decide  
\emph{(i)} whether each party should be assigned to an \emph{existing} expert
or a \emph{new} expert, and  
\emph{(ii)} how to keep the aggregated label distribution within every expert
reasonably balanced.
We cast this assignment step as a facility–location problem that
jointly minimizes \textbf{covariate mismatch},
\textbf{expert-creation-cost}, and
\textbf{label-imbalance}.

\begin{align}
\label{eq:expert_assignment_optimization}
\min_{z, w} \quad &
\sum_{c \in \mathcal{C}} \sum_{k \in \mathcal{K}_0 \cup \mathcal{K}_n} z_{c,k} \cdot \mathrm{MMD}\left(P_t^c(X), P_t^k(X)\right) \nonumber\\
&+\lambda \sum_{k \in \mathcal{K}_n} w_k 
+ \mu \sum_{k \in \mathcal{K}_0 \cup \mathcal{K}_n} \mathrm{JSD}\left(\mathbf{y}_t^k, \bar{\mathbf{y}}_t\right) \nonumber \\
\text{subject to} \quad 
& \sum_{k} z_{c,k} = 1 \quad \forall c \in \mathcal{C} \nonumber \\
& z_{c,k} \le w_k \quad \forall c \in \mathcal{C},\, k \in \mathcal{K}_0 \cup \mathcal{K}_n \nonumber \\
& w_k = 1 \quad \forall k \in \mathcal{K}_0 \nonumber \\
& \sum_{c} z_{c,k} \le U_{\max} \quad \forall k \in \mathcal{K}_0 \cup \mathcal{K}_n
\end{align}

where, $\mathcal{C}$ denotes the set of active parties, with $|\mathcal{C}| = C$ indicating the total number of parties. $\mathcal{K}_0$ and $\mathcal{K}_n$ are the sets of indices corresponding to existing and newly created experts, respectively. The binary variable $z_{c,k} \in \{0,1\}$ indicates whether party $c$ is assigned to expert $k$, while $w_k \in \{0,1\}$ signifies whether expert $k$ is currently active. The parameters $\lambda$ and $\mu$ are hyperparameters that control the trade-off between the cost of training new experts and the penalty for label imbalance across experts.

Each term in the objective has a party-level analogue:
\begin{itemize}
\item \textbf{MMD term} — “Assign me to the expert whose training data looks most like mine.”
\item \textbf{$\lambda$ term} — “Spinning up a fresh model is expensive; do it only when necessary.”
\item \textbf{$\mu$ term} — “Within any expert, keep the label histogram reasonably flat so that future gradient updates are not dominated by a single class.”
\end{itemize}
The constraints ensure every party chooses exactly one expert, new experts cannot be used unless activated, and no expert is overloaded.

The first term minimizes \emph{covariate mismatch} by encouraging parties to be routed to experts whose input distributions closely resemble their own. The second term imposes a flat cost $\lambda$ for each new expert trained, discouraging unnecessary proliferation of models. The third term penalizes \emph{label-distribution divergence} within experts using JSD, promoting balanced class representation in training.

\subsection{Approximate Expert Assignment}

Since the joint optimization problem is NP-hard~\cite{MEGIDDO1982194, FOWLER1981133}, \ours\ uses a modular approximation strategy:
\begin{enumerate}
\item \textbf{Clustering}: Groups parties with similar covariate shifts to minimize MMD mismatch
\item \textbf{Expert Assignment via Latent Memory matching}: Reuses existing experts when possible to control training costs  

\item \textbf{Expert Update using FLIPS/Small Groups}

\item \textbf{New Expert Creation and Training with FLIPS}

%\item \textbf{Label-aware assignment}: Balances label distributions within expert groups to minimize JSD penalty
\item \textbf{Expert Consolidation}: Merges redundant experts to maintain system efficiency
\end{enumerate}

This decomposition enables scalable aggregator-side processing while approximating the optimal joint solution for shift-aware expert management.

%Solving this optimization problem directly is NP-hard []. Instead, \ours\ approximates the solution using a series of modular decisions: (i) latent-representation-based clustering groups parties with similar distributions to minimize MMD-based mismatch; (ii) expert rerouting avoids new expert creation when a matching expert already exists, controlling instantiation cost; (iii) label-based clustering within each cluster aligns with minimizing the JSD penalty; and (iv) consolidation merges redundant experts, reducing overall system complexity. These stages collectively approximate the optimal solution while enabling scalable, adaptive, and efficient model specialization under distributional shifts.

\subsubsection{Clustering}

Once we identify parties exhibiting similar covariate shifts, the central challenge becomes adapting models to remain relevant under the new input distributions. Without adaptation, models risk overfitting to outdated data patterns and underperforming on the shifted distributions. A naive solution would be to treat each shift independently and retrain models from scratch. However, this approach is computationally expensive and fails to exploit structural similarities that often exist across parties’ shifts. Empirically, we observe that many covariate shifts are recurring or follow similar statistical trends, particularly when parties share contextual characteristics (e.g., geography, seasonality, or infrastructure usage patterns).

To leverage this observation, the aggregator clusters parties with similar covariate profiles, and maintains a specialized expert model for each discovered covariate regime. By allowing parties to reuse or fine-tune these experts rather than initiating training from scratch, our approach ensures both efficiency and adaptability. The framework dynamically creates new experts only when a previously unseen covariate regime is encountered, and assigns parties to existing experts when a suitable match is found. This expert-based modularization enables scalable and personalized learning in the presence of recurring and evolving data shifts.

For parties with covariate shift, the aggregator performs K-means clustering on their latent representations $P_t^c(X)$ to group parties with similar distributional changes:
\[
\{G_1, G_2, \ldots, G_k\} \gets \textsc{KMeans}(\mathcal{C}^{\text{cov}}, P_t^c(X))
\]
The optimal number of clusters $k$ is determined using the Davies-Bouldin index~\cite{4766909}. This clustering prevents training redundant experts for parties experiencing similar covariate shifts.
parties whose feature distributions shift in similar directions would need a similar expert model.  Clustering them first and grouping them prevents us from training duplicate experts who learn the same thing.

\subsubsection{Expert Assignment via Latent Memory Matching}

Rather than training new experts for every shift, the aggregator maintains a latent memory, an exponential moving average of each expert’s embedding signatures---to enable expert reuse. For each covariate cluster $G_j$, the aggregator computes the cluster centroid $\bar{P}_j(X)$ and checks for matching existing experts:

\[
\text{If } \min_{k} \mathrm{MMD}(\bar{P}_j(X), \mathcal{M}^{(k)}) \leq \epsilon \text{, then assign } G_j \text{ to expert } k
\]

where $\mathcal{M}^{(k)}$ represents the latent memory signature of expert $k$. This mechanism allows recurring covariate patterns to reuse existing experts without retraining, while preserving privacy since only aggregate embeddings are compared. It also provides a practical approximation to the MMD term in Equation~\ref{eq:expert_assignment_optimization}, avoiding expensive pairwise comparisons across many clients. This process operationalizes the role of the $\lambda$ term in the optimization: $\lambda$ reflects the cost of spawning new experts, but in practice, no manual parameter tuning is required. Instead, we rely on clustering quality metrics, applying the Davies–Bouldin Index with the elbow method to determine when creating additional clusters (and thus new experts) is justified. In this way, the optimization captures the intuition of penalizing unnecessary expert proliferation, and the clustering criteria enforce this trade-off.

\subsubsection{Expert Update using FLIPS/Small Groups}

When a covariate cluster $G_j$ matches an existing expert $\theta^{(k)}$, the system assigns that expert to the cluster. Federated training is then performed using only the matched cohort, keeping expert specialization aligned with the current distribution. To avoid overfitting to skewed or imbalanced label distributions, we employ FLIPS (Federated Learning using Intelligent Participant Selection)~\cite{10.1145/3590140.3629123}, which clusters clients by label proportions and selects participants equitably from each cluster. This ensures that the aggregated training data reflects a balanced label distribution and approximates the $\mu$ term in Equation~\ref{eq:expert_assignment_optimization}. Here, $\mu$ encodes the penalty for label imbalance in the optimization objective; no manual tuning is required: FLIPS directly minimizes Jensen–Shannon Divergence across selected clients, preventing collapse of experts onto dominant classes. Thus, $\mu$ serves to formalize the intuition behind maintaining balanced label coverage, while the FLIPS mechanism realizes this balance automatically during training. For smaller covariate clusters ($|G_j| \leq \gamma$), no federated training is performed; instead, clients perform local finetuning or transfer learning on the assigned expert.

\subsubsection{New Expert Creation and Training with FLIPS} If no existing expert sufficiently matches the latent profile of a covariate cluster $G_j$, the aggregator creates a new expert $\theta^{(j)}$ specialized for this regime. Federated training is then performed using only the selected party cohort $G_j$. To prevent the new expert from overfitting to class-skewed data, we employ FLIPS, which ensures balanced participation by selecting parties with diverse and representative label distributions. This promotes expert specialization while maintaining robustness across the full label space.

%\subsubsection{Expert Update using FLIPS/Small Groups}
\subsubsection{Expert Consolidation} To prevent expert proliferation, the aggregator periodically merges experts with highly similar parameters:
\[
\cos(\theta_i, \theta_j) > \tau \Rightarrow \textsc{MergeExperts}(\theta_i, \theta_j)
\]
where $\cos(\theta_i, \theta_j)$ denotes the cosine similarity between flattened parameter vectors of experts $\theta_i$ and $\theta_j$, and $\tau$ is a similarity threshold. This consolidation step is designed to eliminate redundant or duplicate models that specialize in nearly identical covariate regimes, thereby reducing memory and communication overhead. When two experts are merged, their parameters are averaged in a weighted fashion, and the latent memory is updated accordingly to reflect the new centroid. This maintains a compact, efficient pool of experts while preserving coverage over the full spectrum of observed shifts.

The complete procedure is detailed in Algorithm~\ref{alg:aggregator},~\ref{alg:party}.

\subsection{TEE for Privacy Preservation}
FL ensures raw data never leaves client devices, it still exchanges intermediate artifacts such as model updates, embeddings, and drift statistics, which can leak private information via inference or reconstruction attacks~\cite{membinfsurvey,DLG-git,zhao2020idlg,geiping2020inverting,yin2021see,zhu2019deep}. To strengthen end-to-end privacy, we augment \ours\ with \textbf{Trusted Execution Environments (TEEs)} such as Intel SGX~\cite{intelsgx} or AMD SEV~\cite{amdsev}, which provide hardware-level isolation and verifiable computation. In our design, clients compute $d$-dimensional embeddings locally and transmit them encrypted into a secure enclave, where drift detection (via MMD), clustering, and expert updates are performed without exposure to the aggregator. This prevents leakage of raw data, labels, or model parameters, while enabling secure aggregation of expert updates. TEEs incur modest overhead (e.g., 5\% for AMD SEV)~\cite{10.1145/3590140.3629123} due to enclave initialization and memory constraints, but offer a practical trade-off between privacy and performance. Beyond TEEs, \ours\ shares only aggregate statistics (MMD/JSD scores, label histograms), from which no known reconstruction attacks have been studied. Importantly, TEEs are optional: \ours\ functions without them, but they provide a hardware-assisted layer of privacy protection.

\subsection{\ours\ overheads}

%\ours\ maintains a lightweight memory and computation footprint. On the party side, each device stores a $d$-dimensional feature vector, resulting in $\mathcal{O}(d)$ storage per party. The central aggregator stores expert centroids ($\mathcal{O}(k \cdot d)$), party-to-expert mappings ($\mathcal{O}(n)$), and fixed-size reference data used for MMD-based drift detection. The total aggregator-side space complexity is $\mathcal{O}(k \cdot d + n \cdot d + m \cdot c \cdot h \cdot w)$,, and $c \times h \times w$ are the image dimensions (e.g., $3 \times 224 \times 224$). During operation, MMD drift detection has complexity $\mathcal{O}(m \cdot c \cdot h \cdot w)$, clustering latent representations is $\mathcal{O}(n \cdot d)$, and expert assignment adds $\mathcal{O}(n + k^2)$.
\ours\ maintains a lightweight memory and computation footprint. On the party side, each device stores a single $d$-dimensional feature vector, resulting in $\mathcal{O}(d)$ storage per party. On the aggregator side, memory is required for storing expert centroids ($\mathcal{O}(k \cdot d)$), party-to-expert mappings ($\mathcal{O}(n)$), and a fixed-size reference dataset used for MMD-based drift detection. The total aggregator-side space overhead is $\mathcal{O}(k \cdot d + n \cdot d + m \cdot \mathcal{D})$, where $m$ is the number of reference samples and $\mathcal{D}$ denotes the dimensionality of a single data instance. 
%For example, $\mathcal{D} = c \times h \times w$ for images (e.g., $3 \times 224 \times 224$), $\mathcal{D} = L$ for tokenized text sequences of length $L$, or $\mathcal{D} = C \times T$ for multivariate time-series data with $C$ channels and $T$ time steps. 
During operation, MMD-based drift detection requires $\mathcal{O}(m \cdot \mathcal{D})$ time, clustering latent representations incurs $\mathcal{O}(n \cdot d)$, and expert assignment adds $\mathcal{O}(n + k^2)$. These overheads are lightweight in practice, making \ours\ scalable and applicable across diverse data modalities.

%\pvnote{Should we just say something like $\mathcal{D}$ for data dimension, instead of $c\times\ h \times w $ which is specific to image data? }
%\rbnote{Noted, that makes sense}

\begin{comment}

\subsection{Expert Consolidation and Stability}

To avoid expert proliferation, we periodically merge experts whose weights are highly similar:
\[
\cos(\theta_i, \theta_j) > \tau \Rightarrow \textsc{MergeExperts}(\theta_i, \theta_j)
\]
This step reduces redundancy and consolidates expertise.

\end{comment}

%% file: expertimental_strat.tex
\section{Experimental Strategy}

Our goal is to evaluate the adaptability of \ours\ compared to existing FL techniques under streaming/continual learning conditions, where data distributions shift across windows due to natural or synthetic causes. We focus on the model's ability to maintain high accuracy and fast convergence despite changes in input or label distributions.

\textbf{Datasets:} We evaluate our framework on five datasets spanning diverse covariate and label shifts. FMoW (Functional Map of the World)~\cite{christie2018functionalmapworld} contains over a million satellite images labeled by functional use (e.g., hospital, airport, school) across regions and time; we select 10 labels~\cite{jothimurugesan2023federatedlearningdistributedconcept}. Tiny-ImageNet-C~\cite{hendrycks2019robustness} augments Tiny-ImageNet with 15 corruption types at 5 severities, while CIFAR-10-C~\cite{hendrycks2019robustness} applies the same corruptions to CIFAR-10 for small-scale evaluation. FEMNIST~\cite{leaf-benchmark}, a federated variant of EMNIST, includes handwritten characters from hundreds of users, and Fashion-MNIST~\cite{xiao2017fashionmnistnovelimagedataset} consists of clothing images widely used as a benchmark. We simulate 200 parties for CIFAR-10-C, FEMNIST, and Fashion-MNIST to capture fine-grained heterogeneity in non-IID settings. For FMoW, we instead use 50 parties: its high-resolution imagery and naturally diverse covariates (geography, time, land use) already induce strong variability, so fewer parties suffice while ensuring each has enough samples for stable shift detection. In all cases, 20% of parties are sampled per FL round.

\textbf{Distributional Shifts: } We introduce a range of natural and synthetic covariate and label shifts critical for federated learning evaluation. In FMoW, covariate shifts emerge from visual differences across geographic locations, while label shifts reflect changes in land use distributions~\cite{jothimurugesan2023federatedlearningdistributedconcept}. These represent natural shifts, as we partition the data across different time windows based on temporal metadata. For Tiny-ImageNet-C, we group corruption types and randomly sample severity levels across time windows to simulate unpredictable, environment-driven covariate shifts, providing a robust setting to evaluate \ours. CIFAR-10-C enables similar shift modeling on a smaller scale using controlled corruptions on weather-specific covariate shifts. For FEMNIST and Fashion-MNIST, we simulate synthetic shifts using PyTorch image transformations (e.g., rotation, scaling, color jitter) to induce covariate shifts, and Dirichlet sampling to generate skewed label distributions across time windows. This setup allows us to test both robustness and adaptability under user-specific and temporally evolving distributional dynamics, even in visually simpler tasks. In each window, 50\% of the participating clients retain their previous data distribution, while the remaining 50\% receive a new distribution, simulating partial population shift over time.

\textbf{Models}: We select architectures standard for each dataset: LeNet-5~\cite{lenet} for FEMNIST and FashionMNIST, DenseNet-121~\cite{huang2018densely} for FMoW, ResNet-18~\cite{resnet} for CIFAR-10-C, and ResNet-50~\cite{resnet} for Tiny-ImageNet-C. To detect covariate shifts, we extract encoder-based representations from the penultimate (pre-logit) layer of each model, which captures task-relevant features. Specifically, we use the final average pooling layer in ResNet-18, the global average pooling after the last dense block in DenseNet-121, and the second fully connected layer in LeNet-5. These embeddings are aggregated per party and per window to form latent representations.

\textbf{Comparative Techniques:} To benchmark \ours, we evaluate it against four prominent FL baselines tailored for heterogeneous or streaming scenarios. FedProx~\cite{fedprox} extends FedAvg~\cite{mcmahan2017communication} with a proximal term to stabilize training under non-IID data but relies on a single global model and lacks any mechanism to detect or adapt to evolving distributions over time. OORT~\cite{Oort-osdi21} optimizes party selection using utility-based scores that balance informativeness and system constraints, yet assumes static utility and ignores temporal shifts in data, making it less effective under dynamic conditions. Fielding~\cite{li2024federatedlearningclientsclustering} re-clusters parties based on label distributions to train balanced experts, as in FLIPS~\cite{10.1145/3590140.3629123}, but overlooks covariate shifts and does not adapt clusters as party distributions change across windows. FedDrift~\cite{jothimurugesan2023federatedlearningdistributedconcept} introduces lightweight shift detection by clustering parties based on local loss patterns to assign expert models, but it offers only coarse adaptation and lacks explicit modeling of covariate or label shift dynamics. In contrast, \ours\ explicitly detects both types of shift, dynamically instantiates or reuses experts, and optimizes shift-aware assignment to maintain accuracy and robustness in continual/streaming FL environments.

\textbf{Windowing Strategy.} We use two schemes matched to dataset scale and shift dynamics. For large datasets (FMoW, Tiny-ImageNet-C), we employ \emph{tumbling} windows: each window is disjoint and drawn from distinct covariate distributions—e.g., different timestamps in FMoW and different corruption types in Tiny-ImageNet-C—thereby inducing strong between-window shifts. For smaller datasets (CIFAR-10-C, FEMNIST, Fashion-MNIST), we use \emph{sliding} windows that allow overlap across windows to capture gradual change. FMoW windows align with natural temporal slices, whereas smaller datasets enforce a minimal stable party-size policy to ensure each party retains sufficient data for reliable training.
%This setup better supports gradual and more realistic distributional change in these datasets and allows us to simulate smoother transitions while controlling label and feature distributional shifts explicitly.

\textbf{Metrics Captured:} To evaluate model performance under distributional shifts, we focus on three complementary metrics: Accuracy Drop, Recovery Time, and Max Accuracy. Accuracy Drop measures the immediate decline in performance after a shift, indicating the model’s brittleness to sudden changes—larger drops imply greater sensitivity. Recovery Time captures the number of rounds required to regain 95\% of pre-shift performance, reflecting the model’s adaptability and convergence speed. Max Accuracy denotes the highest accuracy achieved within a window, serving as a measure of the model’s generalization ability under the new distribution. %Together, these metrics offer a comprehensive view of robustness, adaptability, and long-term effectiveness in non-stationary federated learning.

%% file: results_updated.tex
\section{Results and Discussion}
\input{plots_all_new.tex}

\input{tables_all_std.tex}
We evaluate \ours\ on five diverse datasets, covering both large-scale real-world settings (FMoW, TinyImagenet-C) and widely-used academic benchmarks (CIFAR-10-C, FEMNIST, FashionMNIST). As baselines, we compare against four federated learning methods designed to handle party heterogeneity and/or non-stationary data. To simulate streaming and continual learning scenarios, we divide each dataset into multiple time windows where distribution shifts are introduced at fixed intervals. Specifically, we report results over 4 windows for FMoW and CIFAR-10-C, and 5 windows for TinyImagenet-C, FEMNIST, and FashionMNIST. The first window (W0) is used for initialization and burn-in, and all subsequent windows (W1 onward) evaluate adaptation to new, shifted distributions.

Our analysis centers on three key dimensions: (1) \textbf{Adaptation Time}, defined as the number of communication rounds needed for a model to recover after a distribution shift; (2) \textbf{Post-Shift Accuracy}, which captures the final accuracy attained in each window after recovery; and (3) \textbf{System Overheads}, which quantify the additional space and runtime cost of \ours\ compared to other baselines. Specifically, we measure memory usage on both parties and the central aggregator, and latency incurred by shift detection, expert clustering, and expert assignment.

We present the numerical results for Adaptation Time, Accuracy Drop, and Max Accuracy in Tables~\ref{tab:stacked_fmow_cifar} and~\ref{tab:stacked:combined}. We report standard deviations for Accuracy Drop and Max Accuracy across six runs. For Adaptation Time, standard deviations are omitted since the variation is negligible, generally amounting to only a few rounds.  To visualize convergence behavior, we plot Accuracy (\%) over communication rounds in Figures~\ref{fig:fmow:acc} through~\ref{fig:fashion:acc}. Figures~\ref{fig:fmow:maxacc} through~\ref{fig:fashion:maxacc} illustrate the peak accuracy achieved by each method in every window, where error bars represent the variability (standard deviation) observed across six runs. Finally, to illustrate how \ours\ dynamically adapts to distributional changes, we visualize expert assignment distributions across windows in Figures~\ref{fig:fmow:expert:dist} through~\ref{fig:fashion:expert:dist}. These plots show how \ours\ begins with a single expert in W0 and progressively creates and trains new experts as shifts are detected, enabling specialized adaptation to new data regimes.

\textbf{\ours\ reduces adaptation time under distribution shifts.} 
Across all datasets, \ours\ achieves significantly faster adaptation than all baselines. On the large-scale FMoW dataset (Table~\ref{tab:stacked_fmow_cifar}), where most baselines fail to recover within 51 rounds, \ours\ consistently recovers within 1–13 rounds across all windows. In TinyImagenet-C (Table~\ref{tab:stacked:combined}), recovery time drops to 27–29 rounds, whereas other methods plateau with no recovery even after 40 rounds. In CIFAR-10-C, \ours\ completes adaptation in 9–19 rounds, while baselines remain stagnant beyond 51 rounds. On FEMNIST and FashionMNIST, \ours\ achieves recovery within 0–42 rounds, including several windows where adaptation completes in under 5 rounds—markedly outperforming methods that require 50+ rounds without convergence. These trends are also evident in the convergence curves shown in Figures~\ref{fig:fmow:acc} through~\ref{fig:fashion:acc}, where \ours\ consistently demonstrates steeper accuracy recovery slopes and faster stabilization. Even in scenarios where all clients simultaneously experience a shift, clustering still groups similar clients together and triggers creation of new experts, ensuring robustness without collapse.

\textbf{\ours\ consistently achieves the highest post-adaptation accuracy across all datasets and windows}. On top of faster recovery, \ours\ maintains the highest peak accuracy across all settings. In FMoW, it reaches a max accuracy of 66.68\% – 66.14\% across windows, surpassing all other baselines by 15–30 percentage points. In TinyImagenet-C, \ours\ sustains accuracy above 59\% in every window and peaks at 62.15\%, while other methods remain stuck in the 38–54\% range. In CIFAR-10-C, \ours\ achieves 90.86\% in W2 and consistently maintains accuracy above 87\%, which is significantly higher than the 72–78\% ceiling of other approaches. Even on benchmark datasets like FEMNIST and FashionMNIST, \ours\ ends with 66.36\% and 69.00\% respectively, outperforming all baselines that either plateau or decline over time. These trends are visualized in Figures~\ref{fig:fmow:maxacc} through~\ref{fig:fashion:maxacc}, where \ours\ shows both higher peaks and more stable accuracy across windows.

%Interestingly, OORT shows a smaller initial accuracy drop, but this reflects underreaction rather than robustness to shift. Its static party selection masks distribution shifts, leading to limited recovery and lower final accuracy. In contrast, \ours\ detects shifts early and creates and trains new experts, causing a brief dip but enabling rapid recovery and higher long-term performance. These results confirm that \ours\ does not trade off speed for accuracy—it excels in both.
\textbf{Behavior of baselines: }Figures~\ref{fig:fmow:acc} through~\ref{fig:fashion:acc} illustrate accuracy trends across different baselines, and these behaviors can be understood in relation to the types of distribution shifts. FedProx and Fielding rely on a single global model or static label-aware clustering and therefore struggle under covariate shifts, leading to sharp drops and limited recovery. OORT shows a smaller initial accuracy drop, but this reflects underreaction rather than robustness: its utility-based client selection masks distributional changes, preventing effective adaptation and resulting in limited recovery and lower final accuracy. FedDrift performs better under moderate covariate drift, as seen with controlled corrections on CIFAR-10-C, since it clusters clients by loss patterns, but it lacks explicit handling of label shifts and cannot adapt when class distributions skew significantly with the other large-scale datasets like FMoW and Tiny-ImageNet-C. In contrast, \ours\ explicitly detects both covariate and label shifts and instantiates new experts when needed, which explains its accuracy dips followed by rapid recovery and consistently higher post-shift accuracy.

\textbf{\ours\ Overheads:} \ours\ is designed to impose minimal additional memory and computational cost beyond standard FL training, making it practical for real-world deployments. In our ResNet-50 experiments (where a full model is $\approx$100MB), each party stores a single $d$-dimensional latent feature vector (with $d=2048$ for ResNet-50). On the aggregator side, \ours\ maintains: (1) expert centroids (5 experts × 2048 floats = $\approx$40KB); (2) party-to-expert mappings (200 integers = $\approx$0.8KB); and (3) a fixed reference set of 200 RGB images (224×224×3, float32), (4) a group of a maximum of 6 experts = $\approx$600 MB totaling approximately $\approx$714MB.

In terms of runtime, \ours\ adds minimal latency during shift detection and adaptation. Kernel-based MMD drift detection takes an average of 154 ± 17 ms, clustering 200 parties’ latent representations takes 1389 ms, and expert assignment and creation add only 0.15 ms, yielding a total adaptation overhead of approximately 1.55 seconds per shift window. These operations are infrequently triggered, only when a shift is detected, and are amortized over multiple training rounds.

\textbf{Expert Dynamics Reveal Long-Term Adaptation.} To analyze how \ours\ adapts over time, we track the distribution of parties across experts in each window (Figures~\ref{fig:fmow:expert:dist}–\ref{fig:fashion:expert:dist}). At initialization (W0), all parties are assigned to a single expert. This ensures a shared starting point across all methods and datasets, allowing us to isolate and measure adaptation purely as a result of evolving data distributions. From W1 onward, we observe how parties progressively migrate to newly created experts as shifts occur, indicating clear, data-driven reassignments.
In \textbf{FMoW}, parties quickly diversify across multiple experts as new satellite scenes with regional or temporal variations are introduced. By W4, the party population is distributed across five distinct experts, showing that \ours\ recognizes and preserves major distributional changes as separate regimes. In \textbf{TinyImagenet-C}, which contains progressive corruptions (e.g., fog, blur, noise), parties spread smoothly across six experts by W5. The fact that each window results in new dominant experts indicates that \ours\ not only detects shifts but actively expands its capacity to encode new visual distortions without overriding prior knowledge. For \textbf{CIFAR-10-C}, we observe a more contained dynamic: parties gradually shift toward a second expert, stabilizing into a two-expert configuration. This reflects fewer major shifts, and showcases \ours’s ability to remain compact and resist unnecessary expert proliferation when the environment is relatively stable. In \textbf{FEMNIST}, where character distributions and writing styles evolve gradually, parties adapt by migrating across five experts with some reuse over time. This highlights \ours’s ability to handle subtle drift without full resets, preserving older experts when still relevant. \textbf{FashionMNIST} shows mixed behavior, with parties jumping to new experts in W1–W2, re-consolidating around one in W3–W4, and redistributing again in W5. This cyclical pattern reflects changing but partially repeating shifts, demonstrating how \ours\ supports both fresh specialization and expert reuse over time. These evolving expert assignment patterns confirm that \ours\ supports both short-term responsiveness and long-term memory. By allowing parties to switch experts only when necessary and retaining expert states across windows, \ours\ balances adaptation with stability, crucial for robust learning in dynamic federated environments.

%% file: plots_all_new.tex
% --- Row 1: Convergence (FMoW, TinyImageNet-C, CIFAR-10-C) ---
\begin{figure*}[h]
    \centering
    \subfloat[FMoW]{\includegraphics[width=0.32\textwidth]{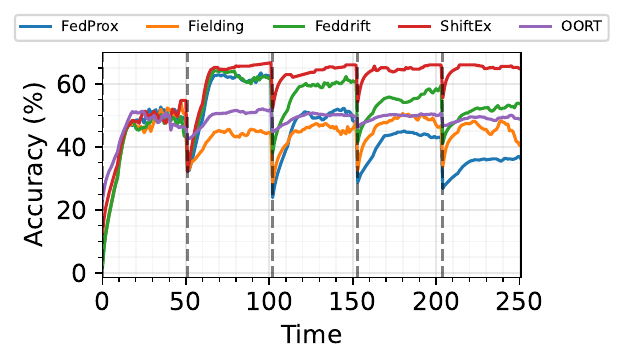}\label{fig:fmow:acc}}\hspace{0pt}
    \subfloat[TinyImageNet-C]{\includegraphics[width=0.32\textwidth]{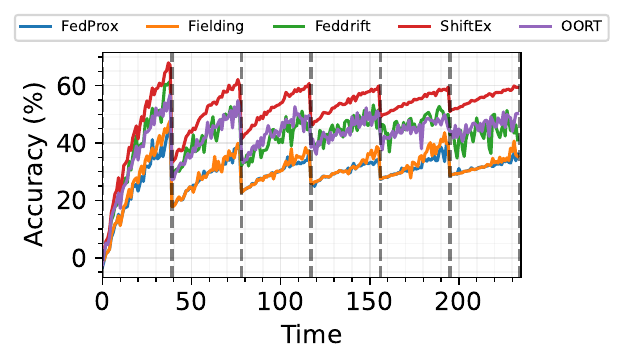}\label{fig:tinyimagenet:acc}}\hspace{0pt}
    \subfloat[CIFAR-10-C]{\includegraphics[width=0.32\textwidth]{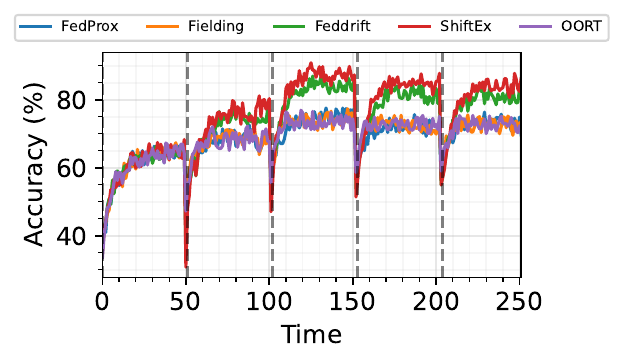}\label{fig:cifar10:acc}}

    \caption{Convergence plots showing test accuracy over rounds for FMoW, TinyImageNet-C, and CIFAR-10-C.}
    \label{fig:convergence:row1}
\end{figure*}

% --- Row 2: Convergence (FEMNIST, FashionMNIST) ---
\begin{figure}[h]
    \centering
    \subfloat[FEMNIST]{\includegraphics[width=0.23\textwidth]{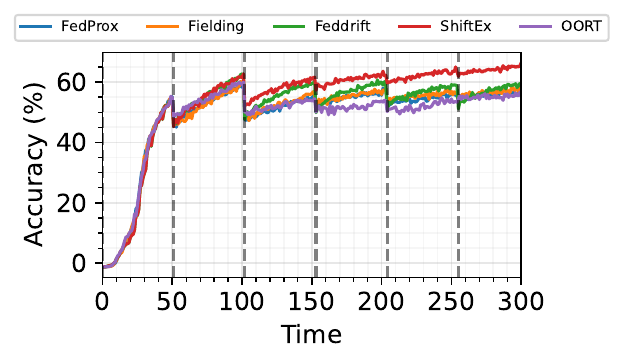}\label{fig:femnist:acc}}\hspace{0pt}
    \subfloat[FashionMNIST]{\includegraphics[width=0.23\textwidth]{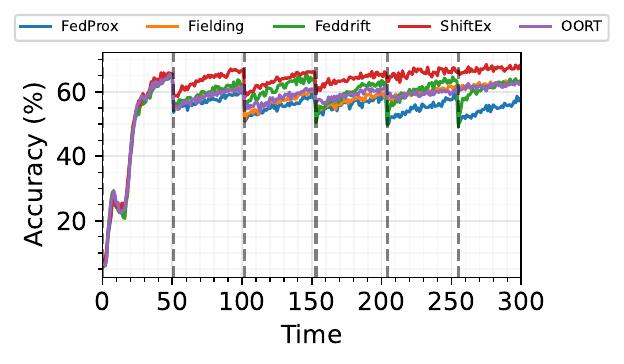}\label{fig:fashion:acc}}

    \caption{Convergence plots showing test accuracy over rounds for FEMNIST and FashionMNIST.}
    \label{fig:convergence:row2}
\end{figure}

% --- Row 1: Max Accuracy (FMoW, TinyImageNet-C, CIFAR-10-C) ---
\begin{figure*}[h]
    \centering
    \subfloat[FMoW]{\includegraphics[width=0.32\textwidth]{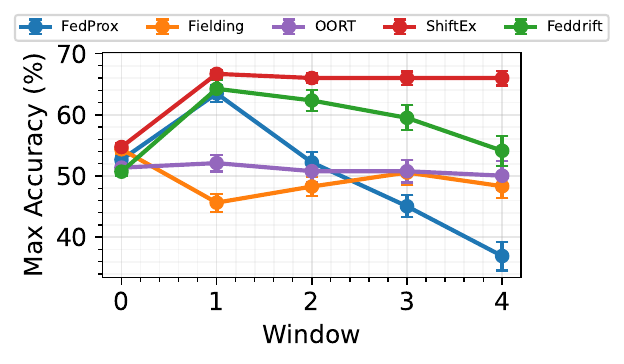}\label{fig:fmow:maxacc}}\hspace{0pt}
    \subfloat[TinyImageNet-C]{\includegraphics[width=0.32\textwidth]{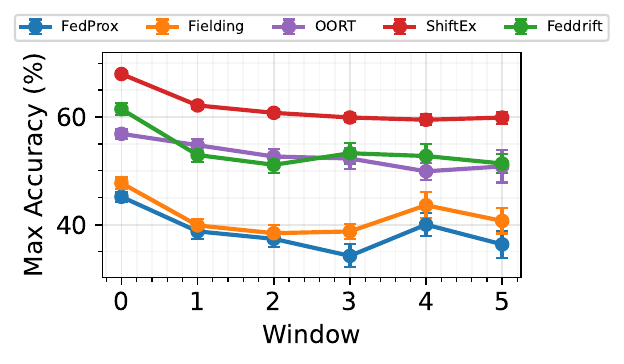}\label{fig:tinyimagenet:maxacc}}\hspace{0pt}
    \subfloat[CIFAR-10-C]{\includegraphics[width=0.32\textwidth]{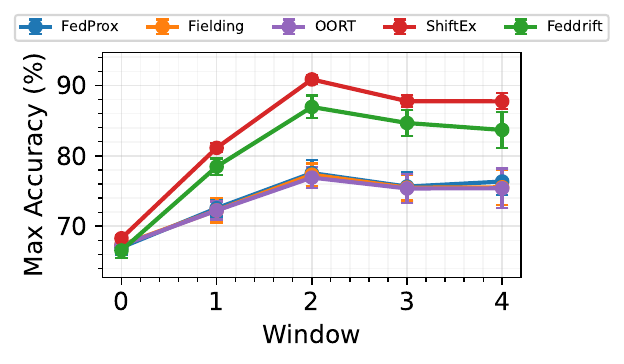}\label{fig:cifar:maxacc}}

    \caption{Maximum accuracy per window for FMoW, TinyImageNet-C, and CIFAR-10-C.}
    \label{fig:maxacc:row1}
\end{figure*}

% --- Row 2: Max Accuracy (FEMNIST, FashionMNIST) ---
\begin{figure}[h]
    \centering
    \subfloat[FEMNIST]{\includegraphics[width=0.23\textwidth]{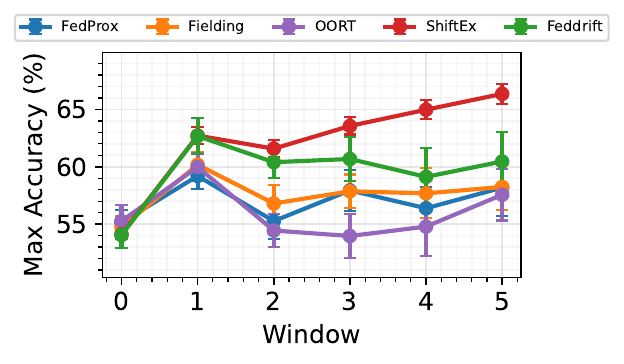}\label{fig:femnist:maxacc}}\hspace{0pt}
    \subfloat[FashionMNIST]{\includegraphics[width=0.23\textwidth]{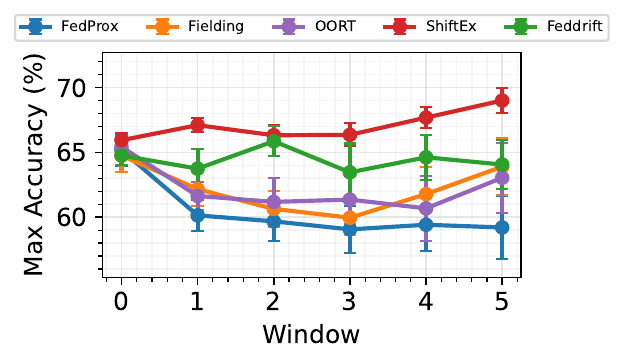}\label{fig:fashion:maxacc}}

    \caption{Maximum accuracy per window for FEMNIST and FashionMNIST.}
    \label{fig:maxacc:row2}
\end{figure}

% --- Row 1: Expert Distribution (FMoW, TinyImageNet-C, CIFAR-10-C) ---
\begin{figure*}[h]
    \centering
    \subfloat[FMoW]{\includegraphics[width=0.25\textwidth]{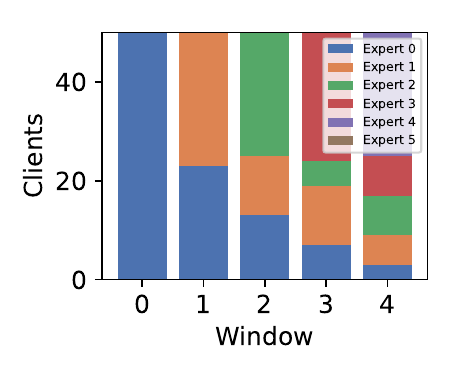}\label{fig:fmow:expert:dist}}\hspace{0pt}
    \subfloat[TinyImageNet-C]{\includegraphics[width=0.25\textwidth]{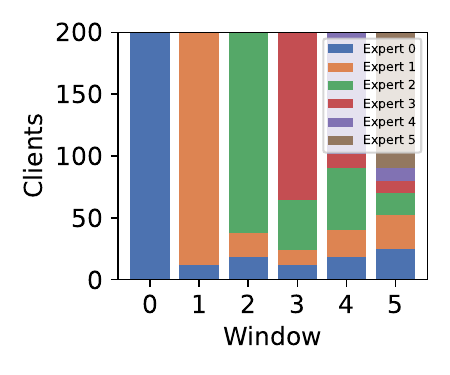}\label{fig:tiny:imagenet:expert:dist}}\hspace{0pt}
    \subfloat[CIFAR-10-C]{\includegraphics[width=0.25\textwidth]{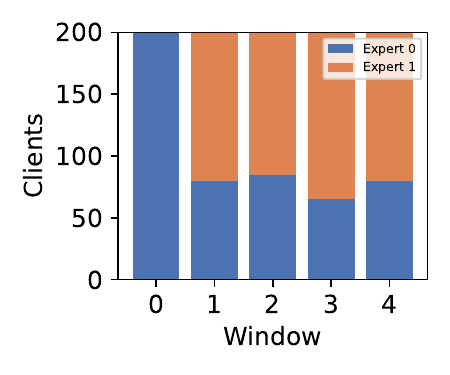}\label{fig:cifar:expert:dist}}

    \caption{Expert distribution across clients for FMoW, TinyImageNet-C, and CIFAR-10-C.}
    \label{fig:expert:dist:row1}
\end{figure*}

% --- Row 2: Expert Distribution (FEMNIST, FashionMNIST) ---
\begin{figure}[h]
    \centering
    \subfloat[FEMNIST]{\includegraphics[width=0.23\textwidth]{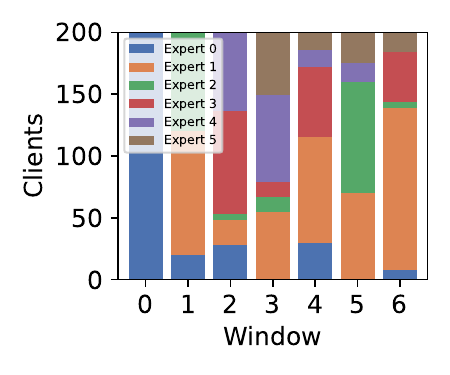}\label{fig:femnist:expert:dist}}\hspace{0pt}
    \subfloat[FashionMNIST]{\includegraphics[width=0.23\textwidth]{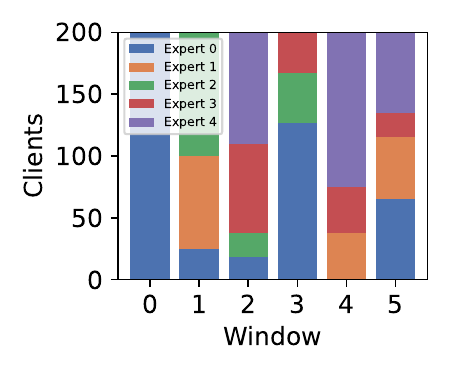}\label{fig:fashion:expert:dist}}

    \caption{Expert distribution across clients for FEMNIST and FashionMNIST.}
    \label{fig:expert:dist:row2}
\end{figure}

%% file: tables_all_std.tex
\begin{table*}[ht]
\centering
% \normalsize
\footnotesize
\setlength{\tabcolsep}{3pt} % tighter column spacing
\begin{tabular}{lcccccccccccc}
\toprule
\multicolumn{13}{c}{\textbf{FMoW}} \\
\toprule
\textbf{Tech.} & \multicolumn{3}{c}{\textbf{W1}} & \multicolumn{3}{c}{\textbf{W2}} & \multicolumn{3}{c}{\textbf{W3}} & \multicolumn{3}{c}{\textbf{W4}} \\
\cmidrule(lr){2-4} \cmidrule(lr){5-7} \cmidrule(lr){8-10} \cmidrule(lr){11-13}
& Drop & Time & Max & Drop & Time & Max & Drop & Time & Max & Drop & Time & Max \\
\midrule
FedProx    & 20.39 $\pm$ 1.36 & 44 & 63.48 $\pm$ 1.38 & 39.57 $\pm$ 1.43 & >51 & 52.23 $\pm$ 1.69 & 23.45 $\pm$ 2.47 & >51 & 45.07 $\pm$ 1.79 & 18.24 $\pm$ 2.51 & >51 & 36.95 $\pm$ 2.51 \\
Fielding  & 21.32 $\pm$ 1.29 & >51 & 45.66 $\pm$ 1.46 & 17.07 $\pm$ 1.83 & >51 & 48.29 $\pm$ 1.57 & 15.76 $\pm$ 1.55 & >51 & 50.57 $\pm$ 2.08 & 16.70 $\pm$ 2.18 & >51 & 48.34 $\pm$ 2.18 \\
OORT      & 8.60 $\pm$ 1.37  & >51 & 52.11 $\pm$ 1.37 & 7.36 $\pm$ 1.39  & >51 & 50.80 $\pm$ 1.85 & 4.68 $\pm$ 2.10  & >51 & 50.79 $\pm$ 2.10 & 3.94 $\pm$ 2.70  & >51 & 50.06 $\pm$ 2.70 \\
\ours\   & 22.57 $\pm$ 0.64 & 13 & 66.68 $\pm$ 0.59 & 14.43 $\pm$ 0.76 & 19 & 66.23 $\pm$ 0.75 & 11.78 $\pm$ 0.91 & 6 & 65.76 $\pm$ 1.07 & 10.49 $\pm$ 1.09 & 4 & 66.14 $\pm$ 1.09 \\
FedDrift  & 18.43 $\pm$ 1.37 & 15 & 64.25 $\pm$ 1.43 & 26.17 $\pm$ 1.63 & >51 & 62.33 $\pm$ 1.70 & 20.84 $\pm$ 2.11 & >51 & 59.50 $\pm$ 2.11 & 18.33 $\pm$ 1.89 & >51 & 54.13 $\pm$ 1.89 \\

\bottomrule
\toprule
\multicolumn{13}{c}{\textbf{CIFAR-10-Corruptions}} \\
\toprule
\textbf{Tech.} & \multicolumn{3}{c}{\textbf{W1}} & \multicolumn{3}{c}{\textbf{W2}} & \multicolumn{3}{c}{\textbf{W3}} & \multicolumn{3}{c}{\textbf{W4}} \\
\cmidrule(lr){2-4} \cmidrule(lr){5-7} \cmidrule(lr){8-10} \cmidrule(lr){11-13}
& Drop & Time & Max & Drop & Time & Max & Drop & Time & Max & Drop & Time & Max \\
\midrule
FedProx    & 17.80 $\pm$ 1.33 & >51 & 72.56 $\pm$ 1.18 & 15.55 $\pm$ 1.48 & >51 & 77.55 $\pm$ 1.81 & 17.01 $\pm$ 2.02 & >51 & 75.64 $\pm$ 2.02 & 18.64 $\pm$ 2.14 & >51 & 76.35 $\pm$ 1.96 \\
Fielding  & 19.76 $\pm$ 1.48 & >51 & 72.25 $\pm$ 1.72 & 16.81 $\pm$ 2.16 & >51 & 77.33 $\pm$ 1.58 & 17.24 $\pm$ 2.03 & >51 & 75.48 $\pm$ 2.03 & 15.39 $\pm$ 1.93 & >51 & 75.54 $\pm$ 2.53 \\
OORT      & 20.03 $\pm$ 1.36 & >51 & 72.24 $\pm$ 1.28 & 16.66 $\pm$ 1.44 & >51 & 76.92 $\pm$ 1.54 & 18.33 $\pm$ 2.06 & >51 & 75.37 $\pm$ 2.06 & 16.39 $\pm$ 2.69 & >51 & 75.39 $\pm$ 2.69 \\
\ours\   & 37.51 $\pm$ 0.53 & >51 & 81.14 $\pm$ 0.65 & 34.07 $\pm$ 0.83 & 9 & 90.86 $\pm$ 0.64 & 39.46 $\pm$ 1.02 & 14 & 87.77 $\pm$ 1.02 & 32.84 $\pm$ 1.09 & 19 & 87.74 $\pm$ 1.09 \\
FedDrift  & 32.16 $\pm$ 1.60 & >51 & 78.47 $\pm$ 1.20 & 27.01 $\pm$ 1.55 & 18 & 86.95 $\pm$ 1.63 & 33.15 $\pm$ 1.83 & 18 & 84.67 $\pm$ 1.83 & 29.73 $\pm$ 1.91 & 23 & 83.68 $\pm$ 1.91 \\

\bottomrule
\end{tabular}
\caption{Comparison of Accuracy Drop (Drop \%), Recovery Time (Time in rounds), and Max Accuracy (Max \%) across Windows 1–4. Top: FMoW dataset. Bottom: CIFAR-10-Corruptions dataset.}
\label{tab:stacked_fmow_cifar}
\end{table*}

\setlength{\tabcolsep}{2pt}
\begin{table*}[ht]
\centering
\scriptsize
\begin{tabular}{lccccccccccccccc}
\toprule
\multicolumn{16}{c}{\textbf{TinyImagenet-C}} \\
\toprule
\textbf{Tech.} & \multicolumn{3}{c}{\textbf{W1}} & \multicolumn{3}{c}{\textbf{W2}} & \multicolumn{3}{c}{\textbf{W3}} & \multicolumn{3}{c}{\textbf{W4}} & \multicolumn{3}{c}{\textbf{W5}} \\
\cmidrule(lr){2-4} \cmidrule(lr){5-7} \cmidrule(lr){8-10} \cmidrule(lr){11-13} \cmidrule(lr){14-16}
 & Drop & Time & Max & Drop & Time & Max & Drop & Time & Max & Drop & Time & Max & Drop & Time & Max \\
\midrule
FedProx    & 27.19 $\pm$ 1.01 & >40 & 38.76 $\pm$ 1.40 & 15.99 $\pm$ 1.53 & >40 & 37.37 $\pm$ 1.44 & 11.49 $\pm$ 1.63 & >40 & 34.21 $\pm$ 2.12 & 6.91 $\pm$ 2.76 & >40 & 40.01 $\pm$ 2.16 & 11.43 $\pm$ 2.39 & >40 & 36.34 $\pm$ 2.57 \\
Fielding  & 29.93 $\pm$ 1.28 & >40 & 39.86 $\pm$ 1.26 & 17.11 $\pm$ 1.40 & >40 & 38.42 $\pm$ 1.52 & 12.40 $\pm$ 1.64 & >40 & 38.75 $\pm$ 1.64 & 11.33 $\pm$ 1.91 & >40 & 43.61 $\pm$ 2.42 & 15.05 $\pm$ 2.05 & >40 & 40.70 $\pm$ 2.05 \\
OORT      & 29.72 $\pm$ 1.13 & >40 & 54.76 $\pm$ 1.18 & 18.87 $\pm$ 1.47 & >40 & 52.66 $\pm$ 1.47 & 14.23 $\pm$ 2.12 & >40 & 52.30 $\pm$ 2.12 & 10.23 $\pm$ 2.21 & >40 & 49.91 $\pm$ 2.56 & 6.01 $\pm$ 2.56 & >40 & 50.83 $\pm$ 2.56 \\
\ours\   & 34.87 $\pm$ 0.52 & 29 & 62.15 $\pm$ 0.61 & 20.73 $\pm$ 0.69 & 30 & 60.78 $\pm$ 0.74 & 14.07 $\pm$ 0.91 & 34 & 59.91 $\pm$ 0.91 & 10.87 $\pm$ 0.99 & 33 & 59.49 $\pm$ 0.99 & 8.53 $\pm$ 1.09 & 30 & 59.89 $\pm$ 1.09 \\
FedDrift  & 33.43 $\pm$ 1.46 & >40 & 52.94 $\pm$ 1.36 & 20.79 $\pm$ 1.49 & >40 & 51.12 $\pm$ 1.49 & 9.76 $\pm$ 1.61 & >40 & 53.27 $\pm$ 1.61 & 10.99 $\pm$ 1.87 & >40 & 52.75 $\pm$ 1.87 & 9.92 $\pm$ 2.21 & >40 & 51.36 $\pm$ 2.21 \\

\bottomrule
\toprule

\multicolumn{16}{c}{\textbf{FEMNIST}} \\
\toprule
    \textbf{Tech.} & \multicolumn{3}{c}{\textbf{W1}} & \multicolumn{3}{c}{\textbf{W2}} & \multicolumn{3}{c}{\textbf{W3}} & \multicolumn{3}{c}{\textbf{W4}} & \multicolumn{3}{c}{\textbf{W5}} \\
    \cmidrule(lr){2-4} \cmidrule(lr){5-7} \cmidrule(lr){8-10} \cmidrule(lr){11-13} \cmidrule(lr){14-16}
     & Drop & Time & Max & Drop & Time & Max & Drop & Time & Max & Drop & Time & Max & Drop & Time & Max \\
    \midrule
FedProx    & 9.36 $\pm$ 1.47 & >51 & 59.20 $\pm$ 1.13 & 11.76 $\pm$ 1.19 & >51 & 55.26 $\pm$ 1.53 & 2.24 $\pm$ 1.69 & >51 & 57.94 $\pm$ 1.81 & 4.02 $\pm$ 1.96 & >51 & 56.38 $\pm$ 1.96 & 1.38 $\pm$ 2.68 & >51 & 58.18 $\pm$ 2.68 \\
Fielding  & 9.14 $\pm$ 1.42 & 48 & 60.18 $\pm$ 1.06 & 11.12 $\pm$ 1.57 & >51 & 56.80 $\pm$ 1.64 & 4.68 $\pm$ 1.92 & >51 & 57.86 $\pm$ 1.92 & 4.72 $\pm$ 2.33 & >51 & 57.68 $\pm$ 2.33 & 3.12 $\pm$ 2.59 & >51 & 58.24 $\pm$ 2.59 \\
OORT      & 6.76 $\pm$ 1.29 & >51 & 59.98 $\pm$ 1.20 & 9.96 $\pm$ 1.45 & >51 & 54.44 $\pm$ 1.40 & 4.40 $\pm$ 1.74 & >51 & 53.96 $\pm$ 1.95 & 3.74 $\pm$ 2.19 & >51 & 54.78 $\pm$ 2.60 & 0.26 $\pm$ 1.89 & >51 & 57.56 $\pm$ 2.31 \\
\ours\   & 8.90 $\pm$ 0.60 & 42 & 62.72 $\pm$ 0.71 & 10.62 $\pm$ 0.75 & 35 & 61.58 $\pm$ 0.73 & 4.32 $\pm$ 0.74 & 14 & 63.56 $\pm$ 0.75 & 3.94 $\pm$ 0.96 & 1 & 64.98 $\pm$ 0.84 & 3.12 $\pm$ 1.02 & 0 & 66.36 $\pm$ 1.02 \\
FedDrift  & 8.14 $\pm$ 1.49 & 40 & 62.68 $\pm$ 1.58 & 15.04 $\pm$ 1.63 & 50 & 60.40 $\pm$ 1.41 & 10.56 $\pm$ 1.72 & 44 & 60.68 $\pm$ 1.72 & 9.82 $\pm$ 2.00 & >51 & 59.12 $\pm$ 2.00 & 8.02 $\pm$ 2.30 & 49 & 60.46 $\pm$ 2.30 \\
    \bottomrule
\toprule
\multicolumn{16}{c}{\textbf{Fashion MNIST}} \\
\toprule
\textbf{Tech.} & \multicolumn{3}{c}{\textbf{W1}} & \multicolumn{3}{c}{\textbf{W2}} & \multicolumn{3}{c}{\textbf{W3}} & \multicolumn{3}{c}{\textbf{W4}} & \multicolumn{3}{c}{\textbf{W5}} \\
\cmidrule(lr){2-4} \cmidrule(lr){5-7} \cmidrule(lr){8-10} \cmidrule(lr){11-13} \cmidrule(lr){14-16}
 & Drop & Time & Max & Drop & Time & Max & Drop & Time & Max & Drop & Time & Max & Drop & Time & Max \\
\midrule
FedProx    & 11.60 $\pm$ 1.15 & 44 & 60.14 $\pm$ 1.19 & 9.28 $\pm$ 1.80 & >51 & 59.68 $\pm$ 1.55 & 5.96 $\pm$ 1.93 & >51 & 59.06 $\pm$ 1.83 & 9.09 $\pm$ 2.32 & >51 & 59.42 $\pm$ 2.32 & 9.33 $\pm$ 2.31 & >51 & 59.21 $\pm$ 2.31 \\
Fielding  & 8.68 $\pm$ 1.34 & 30 & 62.16 $\pm$ 1.29 & 11.08 $\pm$ 1.45 & 47 & 60.64 $\pm$ 1.45 & 5.14 $\pm$ 1.78 & >51 & 59.96 $\pm$ 1.78 & 2.80 $\pm$ 2.41 & 19 & 61.78 $\pm$ 2.41 & 0.34 $\pm$ 2.70 & 0 & 63.90 $\pm$ 2.70 \\
OORT      & 10.10 $\pm$ 1.12 & 33 & 61.64 $\pm$ 1.08 & 6.54 $\pm$ 1.75 & 41 & 61.18 $\pm$ 1.87 & 4.12 $\pm$ 1.78 & 32 & 61.36 $\pm$ 1.78 & 3.26 $\pm$ 1.83 & 19 & 60.68 $\pm$ 2.49 & -0.86 $\pm$ 1.74 & 0 & 63.04 $\pm$ 2.70 \\
\ours\   & 7.56 $\pm$ 0.56 & 5 & 67.10 $\pm$ 0.54 & 8.90 $\pm$ 0.75 & 4 & 66.32 $\pm$ 0.75 & 6.58 $\pm$ 0.74 & 1 & 66.35 $\pm$ 0.89 & 3.24 $\pm$ 0.76 & 0 & 67.68 $\pm$ 0.80 & 1.14 $\pm$ 1.03 & 0 & 69.00 $\pm$ 1.03 \\
FedDrift  & 8.68 $\pm$ 1.29 & 20 & 63.74 $\pm$ 1.51 & 8.09 $\pm$ 1.73 & 13 & 65.86 $\pm$ 1.14 & 15.48 $\pm$ 1.50 & 22 & 63.46 $\pm$ 2.23 & 13.82 $\pm$ 2.50 & 16 & 64.62 $\pm$ 1.88 & 15.64 $\pm$ 2.51 & 10 & 64.06 $\pm$ 2.51 \\
\bottomrule

\end{tabular}
\caption{Accuracy Drop (Drop \%), Recovery Time (Time in rounds), and Max Accuracy (Max \%) across Windows 1–5. Top: TinyImagenet-C. Bottom: Fashion MNIST.}
\label{tab:stacked:combined}
\end{table*}

%% file: relatedworks.tex
\section{Related Work}

FL has emerged as a critical approach to enable privacy-preserving collaborative learning across decentralized data sources. A persistent challenge in FL is handling non-stationary distributions, particularly covariate and label shift.

\textbf{FL under Distributional Shifts.} 
Prior work has studied a range of distributional shifts in FL, including inter- and intra-party, external, prior, test-time, label, and concept shifts. Label shift has been tackled via target-aware aggregation~\cite{zec2024fedpals}, dynamic party selection~\cite{pang2024feddcs}, classifier regularization~\cite{li2021fedrs}, and test-time calibration~\cite{xu2023joint}. Covariate shift has been addressed across multiple axes, including intra-/inter-party~\cite{ramezanikebrya2023federatedlearningcovariateshifts}, external shift~\cite{10.1145/3565010.3569062}, domain-specific shifts~\cite{zhu2025fedweight,xu2023federatedcovariateshiftadaptation}, and test-time adaptation~\cite{tan2023heterogeneity}. Approaches include density ratio matching, contrastive learning~\cite{tan2023heterogeneity}, mixture modeling~\cite{wu2023personalized}, pruning-based regularization~\cite{goksu2024robustfederatedlearningface}, and privacy-preserving transfer~\cite{gao2019privacy}. Catastrophic forgetting and concept drift have also been studied~\cite{huang2022learn,liu2020evaluationframeworklargescalefederated}, and clustering-based adaptations like FedConceptEM~\cite{guo2023fedconceptem} and FlexCFL~\cite{duan2021flexible} group parties by shift type. However, many approaches rely on fixed groupings or prior knowledge. In contrast, \ours{} dynamically detects and responds to label, covariate, and test-time shifts through expert specialization, reuse, and reassignment—enabling continual adaptation without static assumptions.

\textbf{Mixture of Experts in FL.} 
MoE frameworks have shown promise for personalization and heterogeneity in FL. Zec et al.~\cite{zec2020federated,zec2021specializedfederatedlearningusing} first applied MoE to combine local specialists and global generalists. FedMoE~\cite{mei2024fedmoepersonalizedfederatedlearning} and pFedMoE~\cite{yi2024pfedmoe} optimize submodel selection and expert gating for data and model heterogeneity. Other works blend personalized and global outputs~\cite{guo2021pfl}, ensemble party-specific experts~\cite{reisser2021federated}, or share personalized modules~\cite{feng2025pm}. MoE-FL~\cite{parsaeefard2021robustfederatedlearningmixture} enhances robustness via trust-aware aggregation. \ours{} advances this line by making expert assignment shift-aware and temporal: it dynamically spawns, reuses, and merges experts in response to distributional drift, enabling scalable continual adaptation in real-time FL environments.

\textbf{Continual and Temporal Adaptation in FL.} 
To go beyond static training assumptions, recent efforts focus on temporal and streaming FL. Adaptive FL with mixture models~\cite{chen2024adaptive} and time-aware multi-branch networks~\cite{zhu2022diurnal} attempt to model evolving distributions, but rely on fixed priors. Streaming FL techniques~\cite{marfoq2023federatedlearningdatastreams,wang2023local} leverage online data collection and cache update rules to manage distributional drift. Other approaches target evolving concepts through semi-supervised prototypes~\cite{mawuli2023semi}, drift-aware reweighting~\cite{casado2022concept}, and dynamic party clustering~\cite{jothimurugesan2023federatedlearningdistributedconcept,li2024federatedlearningclientsclustering}. 
Federated continual learning~\cite{hamedi2025federatedcontinuallearningconcepts,wang2024federated} and heterogeneity-aware continual learning~\cite{criado2022non} address drift while preserving stability and privacy, while lightweight two-stream architectures~\cite{yao2018two} cut communication under shift. Distinct from these, \ours\ offers a modular MoE framework that continuously adapts to both abrupt and gradual changes without relying on explicit drift signals or fixed temporal structures.

\textbf{Federated continual learning (FCL)}. FCL extends FL to dynamic, non-stationary environments by enabling models to adapt to sequential tasks or streaming data while mitigating catastrophic forgetting. Early work such as FedWeIT \citep{pmlr-v139-yoon21b} proposed task-adaptive parameter decomposition and selective inter-client transfer, while FedCIL \citep{dong2022fcil} introduced prototype-based distillation for class-incremental FL. Rehearsal and regularization approaches, including FedPMR \citep{wang2023fedpmr}, preserve prior knowledge through probability alignment and parameter-consistency constraints. More recently, drift-aware device management for IoT-based FL \citep{zhou2024drift} integrates server-side drift detection with exemplar-based continual learning to balance resource constraints with adaptation. Other federated continual and drift-robust methods include IFCA, which clusters clients into static groups~\citep{ghosh2020ifca}, and test-time adaptation approaches such as FedTHE~\citep{jiang2023fedthe}, ATP~\citep{bao2023atp}, and FedCTTA~\citep{hasan2025fedctta}. Surveys further highlight FCL’s dual challenge of preserving prior knowledge and fusing heterogeneous client experiences \citep{yang2024fclkf,criado2022nonIID}. In contrast, \ours\ contributes a MoE middleware that performs distributional client routing without input-level gating, constrains expert growth via expert consolidation, and narrows the leakage surface by operating only on aggregate latent statistics inside a TEE.

%% file: conclusion.tex
%\section{Towards \ours\ in real deployments}

\section{Conclusion}
We present a shift-aware mixture of experts framework for federated learning under non-stationary data, addressing covariate and label shifts in streaming environments. Through shift detection, latent memory for expert reuse, and facility location-based optimization, our method enables dynamic expert instantiation and scalable adaptation. Extensive experiments show significant gains in accuracy and adaptation speed over FL baselines, offering a practical, privacy-preserving middleware for real-world deployments. Traditional FL middleware approaches assume relatively stable workload characteristics, but our framework addresses the unique challenge of building middleware that can detect, respond to, and manage distributional shifts in real-time across federated participants. The system implements key middleware patterns, including event-driven architectures (responding to detected shifts), load balancing (distributing clients across experts), and service discovery (expert reuse and consolidation). The windowing-based stream processing and latent memory mechanisms function as middleware services that enable scalable, real-time adaptation without requiring application-level awareness of underlying distributional changes. Limitations include potential expert pool growth under extreme heterogeneity, reliance on frozen encoders, and honest client reporting. Under extreme heterogeneity, the number of experts may proliferate; however, our consolidation step periodically merges similar experts to mitigate this risk. Future work will explore expert compression via online distillation, secure reporting under adversarial settings, and tighter integration with hardware enclaves for end-to-end privacy.

% Tasks:
% \begin{itemize}
%     \item Fix references
%     \item Make figures bigger
%     \item Read through it to see if there are any discrepancies
% \end{itemize}